%% file: main.tex
\newif\ifsciencestyle
\sciencestylefalse  %

\ifsciencestyle

\documentclass[12pt]{article}

\usepackage[letterpaper,margin=1in]{geometry}

\renewenvironment{abstract}
	{\quotation}
	{\endquotation}

\date{}

\renewcommand\refname{References and Notes}

\makeatletter
\renewcommand{\fnum@figure}{\textbf{Figure \thefigure}}
\renewcommand{\fnum@table}{\textbf{Table \thetable}}
\makeatother
\usepackage{scicite}
\else
    \documentclass[10pt,journal,compsoc]{IEEEtran}
    \IEEEoverridecommandlockouts
    \usepackage[T1]{fontenc}
    \usepackage{graphicx}
    \usepackage{fix-cm}  %
\fi

\usepackage{graphicx}
\usepackage[final,nopatch=eqnum,nopatch=footnote]{microtype}

\usepackage{url}
\usepackage{hyperref}
\usepackage[frozencache,cachedir=.]{minted}
\usepackage{tikz}
\usepackage{booktabs}
\usepackage{xcolor}
\usepackage{colortbl}
\usepackage{amssymb}
\usepackage{pgfplots}
\usepackage{subcaption}
\pgfplotsset{compat=1.18}
\usepackage{float}
\usepackage{setspace}

\floatstyle{plain}
\restylefloat{codesnippet}

\newfloat{codesnippet}{tbhp}{lop}
\floatname{codesnippet}{Code Snippet}

\def\scititle{
Pixi: Unified Software Development and Distribution for Robotics and AI
}
\title{\bfseries \boldmath \scititle}

\def\sciauthornames{
Tobias Fischer$^{1}$$^\ast$\and Wolf Vollprecht$^{2}$\and Bas Zalmstra$^{2}$\and Ruben Arts$^{2}$\and Tim de Jager$^{2}$\and Alejandro Fontan$^{1}$\and Adam~D~Hines$^{1,3}$\and Michael Milford$^{1}$\and Silvio Traversaro$^{4}$\and Daniel Claes$^{5}$\and Scarlett Raine$^{1}$
}
\def\sciauthorQUT{$^{1}$QUT Centre for Robotics, Queensland University of Technology, 2 George St, Brisbane, QLD 4001, Australia}
\def\sciauthorPrefix{$^{2}$prefix.dev, Hobrechtstraße 58, 12047 Berlin, Germany}
\def\sciauthorMQ{$^{3}$School of Natural Sciences, Macquarie University, Eastern Road, Macquarie Park, NSW 2109, Australia}
\def\sciauthorIIT{$^{4}$Fondazione Istituto Italiano Di Tecnologia (Italian Institute of Technology), Genova, Italy}
\def\sciauthorSmartRobotics{$^{5}$Smart Robotics, De Rijn 6C, 5684 PJ Best, The Netherlands}

\ifsciencestyle
\author
{\sciauthornames\\
\small{\hspace*{-1cm}\sciauthorQUT}\and
\small{\sciauthorPrefix}\and
\small{\hspace*{-0.5cm}\sciauthorMQ}\and
\small{\sciauthorIIT}\and
\small{\sciauthorSmartRobotics}\and
\small{$^\ast$Corresponding author. Email: tobias.fischer@qut.edu.au}
}
\else
    \def\and{, }
    \title{\scititle}
    \author{\sciauthornames
    \thanks{\sciauthorQUT}
    \thanks{\sciauthorMQ}
    \thanks{\sciauthorPrefix}
    \thanks{\sciauthorIIT}
    \thanks{\sciauthorSmartRobotics}
    }
\fi

\begin{document} 

\ifsciencestyle
\else
\IEEEtitleabstractindextext{
\begin{abstract}
\input{abstract}
\end{abstract}
}
\makeatletter
\let\orig@IEEEtitleabstractindextextbox\@IEEEtitleabstractindextextbox
\long\def\@IEEEtitleabstractindextextbox#1{%
  \parbox{%
    \ifCLASSOPTIONcompsoc 0.922\textwidth \else 1\textwidth \fi
  }{%
    \begingroup
      \setlength{\parindent}{0pt}%
      \advance\leftskip by 1.5em\relax
      \advance\rightskip by 1.5em\relax
      #1\par
    \endgroup
  }%
}
\makeatother
\fi

\maketitle 

\ifsciencestyle
\begin{abstract}
\bfseries \boldmath
\input{abstract}
\end{abstract}
\fi

\section*{Introduction}
The robotics revolution promises autonomous systems that will transform society, yet a hidden crisis threatens to slow this progress. Up to 70\% of robotics research software cannot be reproduced by independent teams~\cite{bonneel2020code}, undermining not only real-world deployment, but also the pace of scientific discovery, the credibility of published results, and collaboration across institutions. In robotics and computer vision, these challenges are amplified by the need to integrate multiple programming languages, specialized libraries, hardware-specific drivers, and platform-dependent toolchains. The heterogeneous nature of robotics software stacks—spanning real-time control, perception, planning, and simulation—creates dependency-management challenges that existing tools fail to address effectively.

The development of Pixi reimagines package management for scientific computing, addressing a common pattern in research software: code written to solve immediate problems often becomes unusable over time, while repositories linked to publications are difficult to reproduce by independent teams. Born from the experiences of a team of roboticists and shaped through extensive input from the wider open-source community, Pixi provides a unified solution tailored to the multi-language, cross-platform requirements of modern scientific computing. By abstracting away the software engineering complexities that have become integral to research workflows, it enables scientists to focus on their core scientific questions rather than on environment management.

Traditional package managers evolved from single-language ecosystems, limiting their ability to manage the diverse, multi-language dependencies common in robotics and scientific computing. Pixi, by contrast, was designed from the outset as a multi-language, cross-platform system. It is distributed as a single self-contained binary requiring no base environment or Python installation, allowing straightforward adoption across research settings with a simple one-command installation. The design of Pixi aims to make dependency management an automatic and reliable part of the research workflow, serving as background infrastructure that enables, rather than impedes, scientific collaboration and reproducibility. Through five detailed case studies, spanning academic collaboration, cross-platform MATLAB/C++ workflows, systematic algorithmic comparisons, field-robotics deployment, and industrial production systems, we demonstrate how Pixi integrates seamlessly across the entire robotics development lifecycle.

\begin{figure}
    \centering
    \ifsciencestyle
    \includegraphics[width=0.8\linewidth]{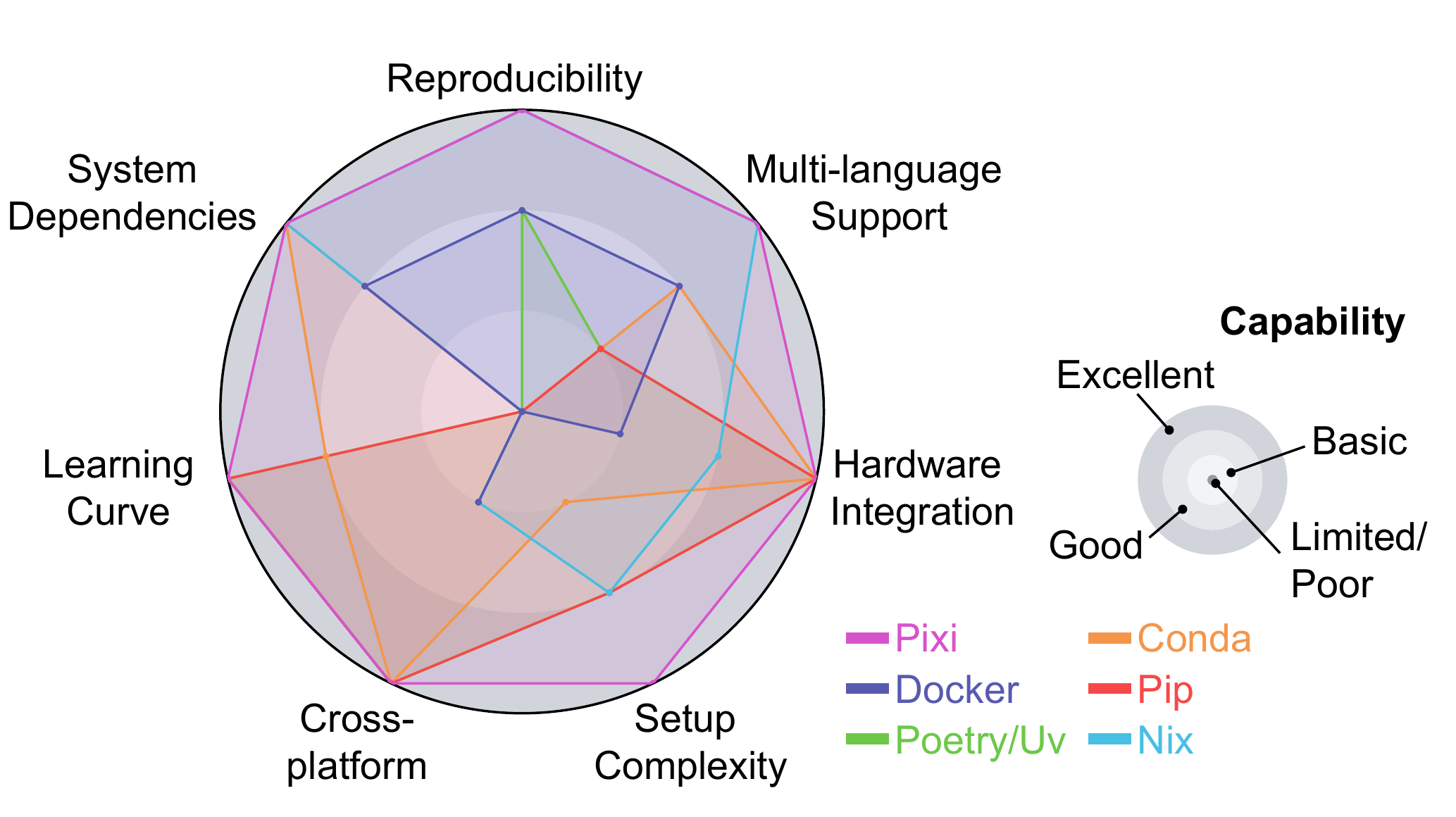}
    \else
    \includegraphics[width=1\linewidth]{Summary_Figure.pdf}
    \caption{Summary of package management approaches for scientific computing and robotics research. 
This figure provides a visual overview of the comparison presented in Table~\ref{tab:package_manager_comparison}.}
    \label{fig:summarypackages}
\end{figure}

Traditional package management approaches suffer from fundamental limitations when applied to scientific computing workflows. Language-specific managers like pip (Python)~\cite{shaffer2021empirical}, npm (JavaScript)~\cite{goswami2020investigating}, or cargo (Rust)~\cite{li2022empirical} cannot handle the multi-language reality of modern robotics systems. System-level package managers such as apt~\cite{krafft2005debian} or homebrew~\cite{jolowicz2024hypermodern} lack reproducibility guarantees and vary dramatically across platforms. Container-based solutions like Docker~\cite{cito2017empirical} provide isolation, ensuring consistent environments across different systems and preventing dependency conflicts, but they introduce significant complexity, performance overhead, and barriers to hardware integration—particularly problematic for robotics applications requiring direct access to sensors, actuators, and specialized compute hardware such as Graphical Processing Units (GPUs). The current conda ecosystem, while addressing some cross-language dependency challenges, introduces its own limitations, including slow dependency resolution, lack of native lockfile support, and complex environment management workflows that burden researchers with system administration tasks rather than enabling focus on scientific innovation. Similarly, tools like Poetry~\cite{poetry} or uv~\cite{uv} that address reproducibility by providing lockfiles for Python-only projects have shown promise but are not suitable for multiple language projects. A comparison of these approaches and their suitability for robotics is provided in Table~\ref{tab:package_manager_comparison} and visualized in Fig.~\ref{fig:summarypackages}.

These limitations have real consequences for scientific progress. Research artifacts become difficult to reproduce, collaboration within and across institutions becomes hindered by environment setup complexity, valuable research time is diverted from scientific inquiry to systems administration, and many researchers avoid releasing code because of concerns about its experimental nature and the significant effort required to make research prototypes suitable for public release.

Pixi overcomes these challenges by providing a unified package-management framework specifically designed to address the reproducibility and collaboration barriers in robotics research. It introduces several key innovations that together transform how researchers manage dependencies and share computational work. First, Pixi implements project-level environments with comprehensive lockfiles that capture exact dependency states across all supported platforms, ensuring that the same environment can be recreated bit-for-bit on any system without the need for containers or manual configuration. Second, it includes a next-generation dependency solver implemented in Rust that uses modern Boolean satisfiability techniques to resolve complex dependency graphs up to 10× faster than conda and about 3× faster than micromamba\footnote{Micromamba is a lightweight, standalone version of the Mamba package manager that serves as a conda alternative}, reducing both computation time and the time researchers spend waiting for environment setup. Third, Pixi builds upon and significantly extends the conda ecosystem, retaining its rich package base while addressing its main limitations: slow solving, lack of native lockfiles, and fragmented workflows. By integrating conda-forge and PyPI under a single interface, Pixi removes the need for researchers to juggle multiple tools or environments.

With Pixi, a project's README can be drastically simplified: for example, simply running \texttt{pixi run start} installs all dependencies on any of the supported operating systems, and executes the main entry point of the project. Beyond addressing technical challenges, Pixi's structured approach may help bridge the gap between research prototyping and public code release by encouraging reproducible workflows from project inception, reducing the overhead that often prevents researchers from sharing their implementations. With Pixi, researchers no longer need to choose between quickly implementable research code that is difficult to release or pre-emptively investing heavy effort into production-ready software; the same reproducible workflow now supports both experimentation and dissemination.

Real-world adoption demonstrates Pixi's transformative impact, with 5,300 active projects\footnote{Obtained from GitHub.com via query \texttt{path:**/pixi.toml} on 25 September 2025.} spanning academic institutions and industry research labs. Most significantly, Pixi has enabled research initiatives previously infeasible under traditional dependency management, including large-scale visual SLAM evaluation frameworks like VSLAM-LAB~\cite{fontan2025vslamlab}, cross-platform reproducible papers such as VPRTempo~\cite{hines2024vprtempo}, hybrid MATLAB/C++ workflows~\cite{bergonti2022modeling}, streamlined field robotics deployments, and industrial ROS-based systems powering over 100 factory robots. Through these detailed case studies spanning the robotics development lifecycle, we demonstrate how unified package management serves as a foundational infrastructure for scientific reproducibility and collaborative research, representing an efficient and broadly enabling approach to addressing the reproducibility crisis in robotics.

\section*{Related Work}
Scientific reproducibility has emerged as a fundamental challenge across computational disciplines, with package management systems playing a critical role in enabling or hindering reproducible research. Here we examine the evolution of reproducibility solutions from academic reporting standards to sophisticated package management systems, highlighting persistent gaps that Pixi addresses.

\subsection*{Reproducibility Challenges in Robotics Research}
The robotics community has systematically documented reproducibility challenges that illuminate broader issues in scientific computing. Bonsignorio established the R-article framework, introducing a three-component publication model requiring supplemental methodology details and complete code repositories~\cite{bonsignorio2017new}. However, this framework primarily addressed documentation standards rather than technical infrastructure for reproducible execution.

Beyond documentation standards, computational workflows in robotics face the additional challenge of \emph{version drift}, where software dependencies evolve independently of research code. Beaulieu-Jones and Greene demonstrated that computational environments can become unreproducible within months without explicit dependency management, with 24\% of workflows failing to execute after one year~\cite{alnoamany2018towards}. This temporal degradation is particularly problematic in robotics, where long development cycles and multi-year projects are common.

Empirical studies further reveal the scope of technical challenges, with Cervera's analysis of papers presented at the 2017 IEEE International Conference on Robotics and Automation finding that code reuse is ``not as straightforward as it could seem,'' with significant effort required to integrate external code into host systems~\cite{cervera2019try}. A follow-up work demonstrated that reproducibility becomes increasingly difficult with software evolution, necessitating not just source code but executable versions with all library dependencies~\cite{cervera2023run}. This follow-up work highlighted the \emph{dependency hell} as a primary barrier, where manual integration creates prohibitive costs for researchers. Dependency hell emerges when robotics projects require multiple libraries with incompatible version constraints, forcing researchers into time-consuming manual resolution of conflicting requirements that often have no satisfactory solution. Pixi's reliance on conda-forge packages, which benefit from extensive migration bots and automated continuous integration infrastructure, significantly alleviates these conflicts by providing consistently maintained and tested package versions with compatible dependency specifications across the ecosystem.

These findings align with broader taxonomies of scientific reproducibility. Benureau and Rougier's influential framework~\cite{benureau2017re} established the ``Five R's'', i.e.~Re-runnable, Repeatable, Reproducible, Reusable, Replicable. This framework identified specific technical barriers including environment dependency, hardware specificity, software evolution, and documentation deficits. Their analysis emphasized that scientific code differs fundamentally from production software in its participation in scientific conclusion formation, requiring specialized approaches to reproducibility.

The Robot Operating System's evolution illustrates further challenges. Macenski et al.'s analysis revealed that ROS package management lacks version locking capabilities, with rosdep handling system-level dependencies without version constraints~\cite{macenski2022robot}. This design creates fundamental reproducibility issues where rebuilding applications at different times may result in different dependency versions, even with identical source code.

Recent advances demonstrate the evolution toward web-based solutions, which provide reproducible execution through standardized browser runtimes that eliminate platform-specific dependencies. For example, the ROS2WASM project eliminates local installation requirements entirely by enabling ROS 2 execution in web browsers through WebAssembly~\cite{fischer2024ros2wasm}. While highly beneficial for accessibility, WebAssembly-based approaches trade execution performance for convenience and remain limited to narrow robotics workflows.

\subsection*{Container Technologies and Environment Isolation}
Container technologies emerged as a leading approach to address reproducibility challenges by providing isolated execution environments. Boettiger identified four major challenges: dependency hell, software environment differences, portability issues, and extension difficulties~\cite{boettiger2015docker}. Docker's layered architecture and scriptable environment description offered significant improvements over virtual machines through reduced overhead and modular reuse.

However, subsequent research revealed fundamental limitations. Moreau et al.'s comprehensive review found that containers face scalability issues in petabyte-scale data processing, security vulnerabilities in shared ecosystems, and performance overhead relative to native execution~\cite{moreau2023containers}. Their review noted that containers do not resolve fundamental issues in dependency management or cross-platform compatibility. Furthermore, workflow management systems like Nextflow have demonstrated the value of containerized approaches for computational reproducibility in bioinformatics, but their focus on containerized execution environments still requires researchers to manage complex workflow definitions and container orchestration rather than addressing fundamental dependency resolution challenges~\cite{di2017nextflow}.

Robotics-specific implementations revealed additional challenges. SwarmRob demonstrated Docker orchestration for multi-robot systems but revealed complexity in networking configuration and hardware abstraction~\cite{portner2018swarmrob}. While containers provide environment isolation, they do not address fundamental issues in dependency resolution algorithms or enable seamless hardware integration required for robotics applications. Real-world deployment experiences, such as the Toyota Human Support Robot Community's containerized development environments across 20+ institutions, demonstrate that even sophisticated container approaches require weeks of coordination overhead for multi-institutional robotics projects, highlighting persistent collaboration challenges~\cite{el2022software}.

\subsection*{Quantifiable Impact of Reproducibility Failures}
Empirical evidence demonstrates the scope of reproducibility challenges across scientific computing domains. Bonneel et al.'s comprehensive analysis of computer graphics research found that only 31.8\% of papers had code that is both available and operational, with significant variation across subfields and time periods~\cite{bonneel2020code}. Critically, their analysis demonstrated a positive correlation between code replicability and citation count, providing empirical evidence that reproducible research achieves greater impact.

These quantitative findings also apply to robotics research. Gym-Ignition demonstrated that standardized robotics simulation platforms could significantly improve reproducibility by providing consistent execution environments~\cite{ferigo2020gym}. However, Gym-Ignition remained domain-specific, addressing robotics simulation rather than general-purpose dependency management.

In the broader AI/ML community, a recent analysis showed that over 70\% of researchers have failed to reproduce another researcher's experiments, with paper submissions to major AI conferences increased by 169\% between 2018 and 2023, while reproducibility rates remain critically low~\cite{desai2025reproducibility,baker2016reproducibility}.

\subsection*{Research Gaps and Opportunities}
This review of related works demonstrates that current approaches to reproducibility in scientific computing remain fragmented and incomplete. While containers provide environment isolation, they do not address fundamental dependency resolution limitations or performance requirements. Existing package managers offer sophisticated algorithms but struggle with cross-platform compatibility, version locking, and the multi-language requirements of robotic systems. Domain-specific solutions address particular use cases but lack the generalizability required for broad scientific adoption in robotics.

Effective reproducibility requires integrated solutions that combine advanced dependency resolution algorithms with comprehensive environment management, cross-platform compatibility, and seamless integration with existing development workflows. The quantitative evidence for reproducibility challenges—31.8\% code availability in graphics~\cite{bonneel2020code}, 35\% configuration issues in Python packages~\cite{peng2023less}, and documented correlation between reproducibility and research impact~\cite{bonneel2020code}—establishes both the scope of the problem and the value of comprehensive solutions. Pixi addresses multiple dimensions of the reproducibility challenge simultaneously: advanced SAT solving for dependency resolution, native cross-platform execution without containerization overhead, and unified management of multi-language scientific software stacks.

\section*{Pixi}
Pixi reimagines package management for scientific computing, designed specifically to address the reproducibility and collaboration challenges endemic to robotics and AI research. Traditional package managers evolved from single-language ecosystems that, while effective within their original scope, were not structured to accommodate the diverse, cross-language requirements that later emerged in scientific computing. Extending them to do so would have required re-engineering core design assumptions and community practices. Pixi instead provides a unified, multi-language, cross-platform solution developed from the outset with these requirements in mind, while maintaining the performance characteristics needed for computationally intensive research workflows.

Pixi acknowledges and is built upon the observations that 1) researchers commonly write code that solves immediate problems but becomes unusable over time, and 2) published code repositories associated with papers often prove difficult to execute, while development workflows require significant effort to transfer code between machines or enable collaboration~\cite{gamblin2015spack,cavalcanti2021software}. A further factor is social rather than technical: within the current academic culture, preparing code for public release is often perceived as additional work with little reward, rather than an integral part of the research process. We hypothesize that a key barrier to public code release is therefore the additional effort required to transform research prototypes into well-documented, platform-independent packages that others can readily use. By encouraging reproducible workflows from project inception, Pixi may help bridge this gap, as researchers who begin new projects using Pixi's structured approach naturally develop code that is already closer to being shareable, potentially reducing the overhead of transitioning from prototype to publishable research artifact. Pixi provides the tools that enable robotics researchers to focus on their core problems rather than software engineering complexities that are currently often too integral to their work.

Pixi is distributed as a single binary that requires no base environment or Python installation, enabling straightforward adoption across research environments. Installation requires only a single command\footnote{Unix: \texttt{curl -fsSL https://pixi.sh/install.sh | bash}\\
Windows: \texttt{iwr -useb https://pixi.sh/install.ps1 | iex}}.%

\subsection*{Core Design and Architecture}
Pixi's architecture implements several design principles that distinguish it from existing package management approaches. Pixi adopts a project-local environment model, eliminating global environment management that is often the source of versioning conflicts and reproducibility issues in traditional systems. Crucially, no root privileges are required, enabling deployment on high-performance computing clusters, shared academic workstations, and containerized environments where users lack administrative access. This design eliminates the need for complex system-level package management while providing seamless collaboration workflows where researchers need only execute \texttt{git clone} followed by \texttt{pixi run start} to reproduce complete computational environments.

The main configuration file specifying dependencies and executable tasks in a project is \texttt{pixi.toml}, which is similar in concept to \texttt{Cargo.toml} or \texttt{pyproject.toml} files in other ecosystems. A simple example can be seen in Code Snippet~\ref{fig:pixi_toml}.

\begin{codesnippet}[t]
\centering
\begin{minted}[fontsize=\small,frame=single,breaklines=true]{yaml}
[workspace]
channels = ["conda-forge"]
name = "hello-world"
platforms = ["linux-64", "osx-arm64", "win-64"]

[tasks]
start = "cowpy hello world"

[dependencies]
cowpy = "1.1.*"
python = "3.13.*"
\end{minted}
\caption{Every Pixi workspace is described by a Pixi manifest file called \texttt{pixi.toml}. This simple example contains a single task \texttt{start} which runs a Python file and two dependencies, \texttt{cowpy} and \texttt{python}.}
\label{fig:pixi_toml}
\end{codesnippet}

The core design addresses five persistent challenges in scientific software management~\cite{cervera2019try,bonneel2020code} reviewed in detail above: (1) inability to reinstall exact software versions after extended periods, (2) environment inconsistencies between local development and continuous integration systems, (3) dependency removal from package repositories causing build failures, (4) semantic versioning violations creating unexpected software breaks, and (5) platform-specific system library differences preventing cross-platform builds.

Pixi's solution centers on comprehensive lockfile management that extends beyond traditional dependency specification. The \texttt{pixi.lock} file is automatically generated from the \texttt{pixi.toml} manifest and kept in sync, providing records of exact package versions, dependency trees, and checksums for all components in the software stack. This approach ensures that environment recreation produces bit-for-bit identical installations, addressing the ``works on my machine'' problem that plagues collaborative scientific research~\cite{meyer2014continuous}. The lock file contains two main sections: First, as shown in Code Snippet~\ref{fig:lock_envs}, the environments that are used in the workspace with their complete set of packages. Second, as shown in Code Snippet~\ref{fig:lock_packages}, the definitions of all packages contained in these environments. While human-readable for inspection, the lock file should not be manually modified.

\begin{codesnippet*}
\centering
\begin{minted}[fontsize=\small,frame=single,breaklines=true,breakafter=/]{yaml}
environments:
  default:
    channels:
      - url: https://conda.anaconda.org/conda-forge/
   packages:
     linux-64:
       ...
       - conda: https://conda.anaconda.org/conda-forge/linux-64/python-3.12.2-hab5b_0_cpython.conda
       ...
     osx-64:
       ...
       - conda: https://conda.anaconda.org/conda-forge/osx-64/python-3.12.2-h9f02_0_cpython.conda
       ...
\end{minted}
\caption{The first part of a \texttt{pixi.lock} file describes the environments -- in this example, there is a single environment called \texttt{default} for the \texttt{linux-64} and \texttt{osx-64} platforms using the \texttt{conda-forge} channel.}
\label{fig:lock_envs}
\end{codesnippet*}

\begin{codesnippet*}
\centering
\begin{minted}[fontsize=\small,frame=single,breaklines=true,breakafter=/]{yaml}
- kind: conda
  name: python
  version: 3.12.2
  build: h9f0c242_0_cpython
  subdir: osx-64
  url: https://conda.anaconda.org/conda-forge/osx-64/python-3.12.2-h9f0c242_0_cpython.conda
  sha256: 7647ac06c3798a182a4bcb1ff58864f1ef81eb3acea6971295304c23e43252fb
  md5: 0179b8007ba008cf5bec11f3b3853902
  depends:
    - bzip2 >=1.0.8,<2.0a0
    - libexpat >=2.5.0,<3.0a0
    - libffi >=3.4,<4.0a0
    - libsqlite >=3.45.1,<4.0a0
    - libzlib >=1.2.13,<1.3.0a0
    - ncurses >=6.4,<7.0a0
    - openssl >=3.2.1,<4.0a0
    - readline >=8.2,<9.0a0
    - tk >=8.6.13,<8.7.0a0
    - tzdata
    - xz >=5.2.6,<6.0a0
  constrains:
    - python_abi 3.12.* *_cp312
  license: Python-2.0
  size: 14596811
  timestamp: 1708118065292
\end{minted}
\caption{The second part of a \texttt{pixi.lock} file describes the packages themselves. This example shows a single package \texttt{python} for the \texttt{osx-64} platform.}
\label{fig:lock_packages}
\end{codesnippet*}

Task automation represents another architectural innovation, replacing platform-specific build and run scripts with declarative task definitions. Complex workflows involving compiler installation, CMake configuration, dataset downloading, and model weight retrieval become reproducible with cross-platform \texttt{pixi tasks} that express dependencies and enable result caching. A multi-language example mixing C++ and Python can be found in Code Snippet~\ref{fig:tasks}.%

\begin{codesnippet}
\centering
\begin{minted}[fontsize=\small,frame=single,breaklines]{yaml}
[tasks]
# Commands as lists so documentation can be added in between.
configure = { cmd = [
    "cmake",
    # Use the cross-platform Ninja generator
    "-G",
    "Ninja",
    # The source is in the root directory
    "-S",
    ".",
    # Build in the .build directory
    "-B",
    ".build",
] }

# Depend on other tasks
build = { cmd = ["ninja", "-C", ".build"], depends-on = ["configure"] }

# Using environment variables
run = "python main.py $PIXI_PROJECT_ROOT"
set = "export VAR=hello && echo $VAR"

# Cross platform file operations
copy = "cp pixi.toml pixi_backup.toml"
clean = "rm pixi_backup.toml"
move = "mv pixi.toml backup.toml"
\end{minted}
\caption{Scientific software development requires comprehensive workflows beyond code execution, including formatting, linting, compilation, testing, and benchmarking. Pixi tasks provide declarative automation for these essential development processes, as illustrated here with a multi-language Python/C++ project.}
\label{fig:tasks}
\end{codesnippet}

\subsubsection*{Positioning Relative to Existing Solutions}

Scientific computing environments require balancing reproducibility with ease of use. Pixi maximizes reproducibility while minimizing setup and maintenance effort. Table~\ref{tab:package_manager_comparison} provides a systematic comparison of package management approaches for scientific computing, with the following analysis of key differences:
\begin{enumerate}
    \item \textbf{Pip, Mamba, and conda:} These popular package managers lack native lockfile functionality. Achieving reproducibility requires manually pinning version numbers or using additional tools like pip-compile—a time-consuming and error-prone process.
    \item \textbf{Poetry and uv:} Poetry and uv provide excellent Python package management with lockfiles but only handle Python packages, while Pixi manages all native packages including Python interpreters, C/C++ compilers, Node.js, and many system-level dependencies.
    \item \textbf{Docker:} While Docker containers provide reproducible environments like Pixi, creating and maintaining Dockerfiles requires significant effort and learning containerization paradigms. Moreover, Dockerfiles often involve multiple package managers (e.g., apt-get and pip), and since neither Dockerfiles nor apt-get understand lockfiles, a Dockerfile alone cannot guarantee reproducibility. Unlike Docker, Pixi operates natively on hardware, enabling full utilization of system capabilities (CUDA, Apple Silicon) and seamless GUI application execution without containerization overhead.
    \item \textbf{Packer:} Can build full VM images but suffers from the same reproducibility issues as Docker with substantially higher complexity.
    \item \textbf{Spack} and \textbf{Nix}: These High-Performance-Computing-focused package managers offer sophisticated approaches to dependency management, however their trade-offs do not align with the requirements of robotics research. Spack's~\cite{gamblin2015spack} source-based builds enable precise optimization control but introduce prohibitive compilation times for iterative development. A typical robotics workspace might require 2 to 6 hours to compile from source what Pixi installs in minutes through binary packages. Nix~\cite{bzeznik2017nix} provides mathematical reproducibility guarantees through its functional approach but requires learning functional programming concepts and Nix expression language—cognitive overhead that diverts researcher attention from scientific problems. Pixi combines Nix-like reproducibility guarantees with familiar imperative syntax and conda's binary ecosystem, optimizing for researcher productivity rather than system administrator control.
    \item \textbf{Bazel:} Bazel is a high-performance build system rather than just a package manager. It ensures reproducible and hermetic builds by compiling code with well-defined build rules, which makes it attractive for large-scale software engineering projects. However, Bazel does not resolve or install external dependencies from ecosystems such as conda-forge or PyPI, leaving environment provisioning to other tools. It also has a steep learning curve and high maintenance overhead, making it less suitable for fast-moving research projects. This makes Bazel complementary to Pixi: Pixi manages and locks down the software environment across languages and platforms, while Bazel can be layered on top to provide deterministic builds of the project itself.
\end{enumerate}

\begin{table*}[t]
\centering
\caption{Comparison of package management approaches for scientific computing and robotics research. Evaluation criteria focus on capabilities essential for reproducible, collaborative robotics workflows.}
\label{tab:package_manager_comparison}
\resizebox{\textwidth}{!}{%
\begin{tabular}{@{}l@{\hspace{6pt}}c@{\hspace{6pt}}c@{\hspace{6pt}}c@{\hspace{6pt}}c@{\hspace{6pt}}c@{\hspace{6pt}}c@{}}
\toprule
\textbf{Capability} & \textbf{Pixi} & \textbf{Docker} & \textbf{Poetry/Uv} & \textbf{Conda/Mamba} & \textbf{Pip} & \textbf{Nix} \\
\midrule
\addlinespace[2pt]
\textbf{Reproducibility} & \cellcolor{green!20}{\Large\checkmark\checkmark\checkmark} & \cellcolor{yellow!20}{\Large\checkmark\checkmark} & \cellcolor{yellow!20}{\Large\checkmark\checkmark} & \cellcolor{red!20}{\Large\texttimes} & \cellcolor{red!20}{\Large\texttimes} & \cellcolor{green!20}{\Large\checkmark\checkmark\checkmark} \\
\addlinespace[1pt]
\textit{Native lockfiles} & \textit{Yes} & \textit{Container isolation} & \textit{Python only} & \textit{Manual pinning} & \textit{Manual pinning} & \textit{Functional builds} \\
\addlinespace[4pt]

\textbf{Multi-language Support} & \cellcolor{green!20}{\Large\checkmark\checkmark\checkmark} & \cellcolor{yellow!20}{\Large\checkmark\checkmark} & \cellcolor{yellow!20}{\Large\checkmark} & \cellcolor{yellow!20}{\Large\checkmark\checkmark} & \cellcolor{yellow!20}{\Large\checkmark} & \cellcolor{green!20}{\Large\checkmark\checkmark\checkmark} \\
\addlinespace[1pt]
\textit{C++, Python, Rust, etc.} & \textit{All languages} & \textit{Via containers} & \textit{Python + C extensions} & \textit{Most languages} & \textit{Python + C extensions} & \textit{All languages} \\
\addlinespace[4pt]

\textbf{Hardware Integration} & \cellcolor{green!20}{\Large\checkmark\checkmark\checkmark} & \cellcolor{yellow!20}{\Large\checkmark} & \cellcolor{green!20}{\Large\checkmark\checkmark\checkmark} & \cellcolor{green!20}{\Large\checkmark\checkmark\checkmark} & \cellcolor{green!20}{\Large\checkmark\checkmark\checkmark} & \cellcolor{yellow!20}{\Large\checkmark\checkmark} \\
\addlinespace[1pt]
\textit{CUDA, Apple Silicon, etc.} & \textit{Native access} & \textit{Basic support} & \textit{Native access} & \textit{Native access} & \textit{Native access} & \textit{Slow CUDA adoption} \\
\addlinespace[4pt]

\textbf{Setup Complexity} & \cellcolor{green!20}{\Large\checkmark\checkmark\checkmark} & \cellcolor{red!20}{\Large\texttimes} & \cellcolor{yellow!20}{\Large\checkmark\checkmark} & \cellcolor{yellow!20}{\Large\checkmark} & \cellcolor{yellow!20}{\Large\checkmark\checkmark} & \cellcolor{yellow!20}{\Large\checkmark\checkmark} \\
\addlinespace[1pt]
\textit{Single command install} & \textit{One command} & \textit{Dockerfile creation} & \textit{Python projects} & \textit{Environment setup} & \textit{Simple install} & \textit{Simple with flakes} \\
\addlinespace[4pt]

\textbf{Cross-platform} & \cellcolor{green!20}{\Large\checkmark\checkmark\checkmark} & \cellcolor{yellow!20}{\Large\checkmark} & \cellcolor{green!20}{\Large\checkmark\checkmark\checkmark} & \cellcolor{green!20}{\Large\checkmark\checkmark\checkmark} & \cellcolor{green!20}{\Large\checkmark\checkmark\checkmark} & \cellcolor{yellow!20}{\Large\checkmark} \\
\addlinespace[1pt]
\textit{Linux, macOS, Windows} & \textit{All platforms} & \textit{Platform dependent} & \textit{All platforms} & \textit{All platforms} & \textit{All platforms} & \textit{No Windows} \\
\addlinespace[4pt]

\textbf{Learning Curve} & \cellcolor{green!20}{\Large\checkmark\checkmark\checkmark} & \cellcolor{red!20}{\Large\texttimes} & \cellcolor{green!20}{\Large\checkmark\checkmark\checkmark} & \cellcolor{yellow!20}{\Large\checkmark\checkmark} & \cellcolor{green!20}{\Large\checkmark\checkmark\checkmark} & \cellcolor{red!20}{\Large\texttimes} \\
\addlinespace[1pt]
\textit{Researcher accessibility} & \textit{Minimal} & \textit{High (containers)} & \textit{Minimal} & \textit{Moderate} & \textit{Minimal} & \textit{Complex language} \\
\addlinespace[4pt]

\textbf{System Dependencies} & \cellcolor{green!20}{\Large\checkmark\checkmark\checkmark} & \cellcolor{yellow!20}{\Large\checkmark\checkmark} & \cellcolor{red!20}{\Large\texttimes} & \cellcolor{green!20}{\Large\checkmark\checkmark\checkmark} & \cellcolor{red!20}{\Large\texttimes} & \cellcolor{green!20}{\Large\checkmark\checkmark\checkmark} \\
\addlinespace[1pt]
\textit{Compilers, libraries, etc.} & \textit{Fully managed} & \textit{Via base images} & \textit{Not handled} & \textit{Fully managed} & \textit{Not handled} & \textit{OS-level management} \\
\addlinespace[2pt]

\bottomrule
\end{tabular}
}
\footnotesize
\\\vspace*{3mm}
\textbf{Symbols:} {\checkmark\checkmark\checkmark} Excellent, {\checkmark\checkmark} Good, {\checkmark} Basic, {\texttimes} Limited/Poor \\
\textbf{Note:} Evaluation based on capabilities relevant to multi-language robotics research workflows requiring reproducibility, hardware integration, and cross-platform collaboration (see also the visual summary in Fig.~\ref{fig:summarypackages}).
\end{table*}

\subsection*{Performance and Scalability Innovations}

Scientific computing workflows demand both correctness and performance from package management systems. Pixi addresses performance limitations through two major technical innovations that achieve substantial speed improvements over existing solutions (see Supplementary Materials for detailed performance analysis). First, a new dependency solver implemented in Rust incorporates modern Boolean satisfiability problem (SAT) solving techniques including conflict-driven clause learning and 2-literal watching, achieving 3× speedup compared to micromamba and 10× improvement over conda for typical scientific workloads. Second, a sharded repository infrastructure replaces monolithic repodata files with content-addressable storage, enabling up to 50× performance improvements for complex environment resolution while supporting immediate package updates essential for active research projects. These innovations collectively transform package management from a workflow bottleneck into transparent infrastructure that supports rapid scientific development cycles.

\subsection*{Scientific Ecosystem Integration}
Pixi is designed as a comprehensive solution that complements and enhances existing scientific workflows rather than replacing them. Its integration capabilities address common pain points in scientific software development while preserving familiar toolchains.

\textbf{Version Control Integration:} Pixi seamlessly integrates with Git-based workflows by treating the \texttt{pixi.lock} file as a version-controlled artifact alongside source code. This ensures that environment specifications evolve synchronously with code changes, enabling researchers to reproduce exact computational environments from any commit in their project history.

\textbf{Computational Notebook Support:} Jupyter and JupyterLab integration exemplifies Pixi's ecosystem compatibility, where notebooks become truly reproducible artifacts. By specifying Jupyter as a project dependency, teams ensure that all collaborators share identical package versions and dependencies, eliminating the common problem where notebook sharing fails due to environment inconsistencies.

\textbf{Multi-Language Project Management:} Unlike traditional virtual environments that focus on single-language ecosystems, Pixi manages complete development stacks including Python interpreters, C/C++ compilers, system libraries, and domain-specific tools within unified project environments. This comprehensive approach proves particularly valuable for robotics projects that integrate algorithms across multiple programming languages and hardware platforms.

\subsubsection*{Multi-Ecosystem Package Foundation}
Pixi's effectiveness derives from its deep integration with established scientific computing ecosystems, particularly conda-forge and PyPI. Conda-forge provides an ideal foundation through its community-driven collection of recipes, reproducible build infrastructure, and policy of never deleting packages, ensuring long-term reproducibility essential for scientific research. The ecosystem's binary package distribution model proves particularly valuable for scientific computing, enabling researchers to focus on problems rather than build system management, while also ensuring consistent compilation across installations.

However, modern scientific computing increasingly requires integration across multiple package ecosystems. While conda-forge provides excellent coverage for scientific libraries, certain packages remain exclusively available through PyPI or other repositories. Pixi addresses this limitation through native PyPI integration via the uv resolver, enabling unified management and consistent dependency solving across both ecosystems. Lockfiles capture exact versions and checksums for both conda and PyPI packages, ensuring complete reproducibility of mixed-ecosystem environments.

This multi-ecosystem approach proves particularly valuable for robotics research where essential libraries from conda-forge must coexist with specialized research packages available only through PyPI, while maintaining the performance and multi-language support (C++, Python, domain-specific languages) required for complex robotics software stacks spanning ROS packages, computer vision libraries, and deep learning frameworks.

\subsection*{Domain-Specific Impact: Robotics and AI}
Robotics research presents unique package management challenges that existing tools struggle to address comprehensively. Unlike domains that rely on single-language applications served by dedicated package managers (cargo, pip, npm), robotics requires coordinating libraries across multiple languages: C++ for real-time control, Python for high-level algorithms, and specialized domain-specific tools. Additionally, modern robotics and AI research depends heavily on hardware acceleration through CUDA, AVX-512, and other optimizations for real-time performance and large-scale computation.

Pixi addresses these interconnected challenges through unified multi-language management and hardware-aware optimization. For example, a single project configuration can specify ROS packages, hardware-optimized OpenCV builds, CUDA-enabled PyTorch installations, compiler toolchains, and development utilities. The binary distribution model inherited from conda eliminates lengthy compilation procedures, enabling rapid collaboration onboarding. Thanks to NVIDIA's contributions to conda-forge\footnote{\url{https://github.com/conda-forge/conda-forge.github.io/issues/1963}} and the system-requirements feature, hardware-optimized package variants are automatically selected, ensuring installations automatically leverage available instruction sets and specialized hardware.

This approach proves particularly valuable where performance directly impacts system capabilities: real-time perception algorithms benefit from optimized computer vision libraries, while large-scale simulations leverage GPU acceleration seamlessly, all managed through a unified interface.

\section*{Enabling Package Creation: Pixi-build and rattler-build}
While Pixi addresses the consumption and management of existing packages, the complete reproducibility ecosystem requires robust tools for package creation and distribution. Scientific research frequently produces novel algorithms, modified versions of existing libraries, and domain-specific tools that must be packaged for community sharing and long-term reproducibility. Robotics research exemplifies this need: perception algorithms require custom builds of computer vision libraries, control systems demand real-time optimized packages, and research frameworks need specialized configurations, unavailable in standard distributions. Pixi-build and rattler-build complement Pixi's consumption-focused tools by enabling package creation and distribution. While rattler-build serves as a standalone command-line tool for creating conda packages from recipe specifications, pixi-build offers specialized backend implementations that enable building conda packages directly from source code across multiple programming languages and build systems.

The traditional conda package building process has presented significant barriers to scientific software developers. Complex build recipes, platform-specific inconsistencies, lengthy compilation times, and debugging difficulties have discouraged researchers from contributing their tools to the broader community. These limitations directly impact scientific reproducibility by creating a gap between research code and community-accessible packages.

Pixi-build addresses these challenges through a comprehensive reimplementation of conda package building infrastructure in Rust, designed specifically to meet the performance, reliability, and usability requirements of scientific software development. The Rust foundation enables static typing throughout the codebase, provides comprehensive documentation, and delivers significantly improved performance compared to traditional Python-based build tools. This implementation provides several key advantages: static typing enhances maintainability and reduces build script errors, native async capabilities enable parallel processing of build stages to significantly reduce compilation times for complex scientific packages, and comprehensive error reporting and debugging support help researchers efficiently diagnose build failures.

\subsection*{Technical Architecture and Design}
Both pixi-build and rattler-build create standard conda packages that integrate seamlessly with the broader conda ecosystem, and existing package management tools. A conda package is fundamentally a compressed archive (\texttt{.tar.bz2} or \texttt{.conda}) containing files that are extracted into standardized installation directories such as \texttt{\$PREFIX/lib}, \texttt{\$PREFIX/share}, and \texttt{\$PREFIX/bin}. Rattler-build creates these cross-platform relocatable packages from simple recipe specifications (see Code Snippets~\ref{fig:evo_recipe} and~\ref{fig:zenoh_recipe} for examples), with a format heavily inspired by conda-build. The Conda Recipe Manager (CRM) automates the conversion process from traditional conda-build formats to rattler-build specifications, providing tools for parsing, rendering, and editing existing recipe files that facilitate migration to the new build system. The output packages maintain full compatibility with existing conda ecosystem tools, including conda, mamba, and pixi.

Rattler-build is a command-line tool that implements a modern approach to cross-platform package building, which addresses fundamental limitations of existing tools. It processes declarative recipe files that specify package metadata, source locations, build procedures, and testing requirements, then orchestrates the complete package creation workflow across multiple platforms and architectures, to produce standard \texttt{.conda} packages compatible with the broader ecosystem.

The build process follows a seven-stage pipeline designed for reproducibility and debugging efficiency:
\begin{enumerate}
    \item \textbf{Recipe rendering:} Parse recipe files and evaluate conditional expressions, Jinja templating, and build variants that generate platform-specific build configurations.
    \item \textbf{Source acquisition:} Retrieve source materials from diverse locations including version control repositories, archive files, and local directories, applying specified patches and verification procedures.
    \item \textbf{Environment preparation:} Establish isolated build and host environments with precise dependency specifications, ensuring consistent compilation contexts across platforms.
    \item \textbf{Build execution:} Run platform-specific build scripts within prepared environments, capturing build outputs and intermediate artifacts for debugging purposes.
    \item \textbf{Binary preparation:} Process compiled outputs to ensure relocatability, including runtime path modification on Unix systems and dependency embedding for cross-platform compatibility.
    \item \textbf{Package assembly:} Bundle processed files with comprehensive metadata including dependency specifications, file manifests, and verification checksums into standard conda package formats.
    \item \textbf{Quality assurance:} Execute specified test suites within clean environments to verify package functionality before distribution.
\end{enumerate}

Recipe specifications follow a structured format that balances expressiveness with simplicity, as summarized in Table~\ref{tab:recipe_sections}. This approach enables researchers to specify complex build requirements while maintaining readability and version control compatibility, essential for collaborative development. Exemplary recipes are shown in Code Snippets~\ref{fig:evo_recipe} and~\ref{fig:zenoh_recipe}.

\begin{table*}[t]
\centering
\caption{Rattler-build recipe sections and their functions in scientific package development}
\label{tab:recipe_sections}
\begin{tabular}{|l|p{0.65\textwidth}|}
\hline
\textbf{Recipe section} & \textbf{Description} \\
\hline
\texttt{context} & Defines template variables for recipe parameterization, enabling reusable build specifications across package variants \\
\hline
\texttt{package} & Specifies package identity including name, version, and metadata required for dependency resolution \\
\hline
\texttt{source} & Defines source code locations, retrieval methods, and integrity verification procedures \\
\hline
\texttt{build} & Configures build environment settings, compilation scripts, and platform-specific optimizations \\
\hline
\texttt{requirements} & Specifies build-time, runtime, and testing dependencies with precise version constraints \\
\hline
\end{tabular}
\end{table*}

\begin{codesnippet}
\centering
\begin{minted}[fontsize=\small,frame=single,breaklines]{yaml}
context:
  name: evo
  version: '1.31.1'

package:
  name: ${{ name|lower }}
  version: ${{ version }}

source:
  url: https://pypi.org/packages/source/e/
       evo/evo-${{ version }}.tar.gz
  sha256: 81bd49dbdc...

build:
  noarch: python
  script: ${{ PYTHON }} -m pip install . --no-deps --no-build-isolation
  python:
    entry_points:
    - evo = evo.main_evo:main
    - evo_traj = evo.entry_points:traj

requirements:
  host:
  - python >=3.10
  - pip
  run:
  - python >=3.10
  - matplotlib-base
  - numpy
  - pandas
  - scipy
  - rosbags >=0.9.20

tests:
  python:
    imports:
    - evo
    pip_check: true

about:
  homepage: https://github.com/MichaelGrupp/evo
  summary: Python package for the evaluation of odometry and SLAM
  license: GPL-3.0-or-later
\end{minted}
\caption{Example rattler-build recipe for evo, a robotics SLAM evaluation package.}
\label{fig:evo_recipe}
\end{codesnippet}

\begin{codesnippet}
\centering
\begin{minted}[fontsize=\small,frame=single,breaklines=true,breakafter=/]{yaml}
context:
  name: zenoh-pico
  version: '1.4.0'

package:
  name: ${{ name|lower }}
  version: ${{ version }}

source:
  url: https://github.com/eclipse-zenoh/
       ${{ name }}/archive/refs/tags/
       ${{ version }}.tar.gz
  sha256: 9cbe440cc1de...

build:
  number: 0
  script:
    - cmake -S . -B build ${CMAKE_ARGS}
    - cmake --build build
    - cmake --install build

requirements:
  build:
    - ${{ compiler('c') }}
    - ${{ compiler('cxx') }}
    - cmake
    - make

tests:
  package_contents:
    lib:
      - libzenohpico
    include:
      - zenoh-pico.h

about:
  homepage: https://zenoh.io
  summary: Lightweight C implementation of the Zenoh protocol for constrained devices
  license: EPL-2.0 OR Apache-2.0
\end{minted}
\caption{Example rattler-build recipe for zenoh-pico, demonstrating C/C++ library packaging with CMake build system for robotics communication protocols. Note that the same recipe is used to build Linux, macOS and Windows conda packages.}
\label{fig:zenoh_recipe}
\end{codesnippet}

\subsection*{Pixi-Build: Language-Specific Backend Architecture}
While rattler-build requires researchers to write explicit recipe files specifying how to build their packages, pixi-build takes a different approach by automatically generating conda packages directly from source code. Pixi-build achieves this through specialized backend programs—each designed for specific programming languages like Python, Rust, or ROS, that automatically detect project structure, dependencies, and build requirements without requiring manual recipe creation.

This backend system offers several advantages. Researchers can create distributable packages without learning recipe syntax; each backend understands the conventions and tools specific to its language; new backends can be added independently without modifying Pixi's core; and all backends produce standard conda packages compatible with existing tools.

Currently available backends span diverse scientific computing domains: \texttt{pixi-build-cmake} for C/C++ projects using CMake, \texttt{pixi-build-python} for PEP 517/518-compliant Python projects, \texttt{pixi-build-rattler-build} for direct recipe.yaml builds with full control, \texttt{pixi-build-ros} for ROS packages with automatic dependency mapping, \texttt{pixi-build-rust} for Cargo-based Rust applications, and \texttt{pixi-build-mojo} for Mojo applications. All backends maintain cross-platform compatibility across Linux, macOS, and Windows, while integrating seamlessly with the broader conda ecosystem.

The ROS backend exemplifies the system's capability to handle domain-specific requirements. It automatically reads package metadata from \texttt{package.xml} files, supports both ROS1 and ROS2 distributions, handles multiple build systems (ament\_cmake, ament\_python, catkin, and cmake), and provides automatic dependency mapping using RoboStack's conda package mappings. This eliminates the traditional complexity of ROS package building while maintaining full compatibility with existing ROS development workflows.

\begin{codesnippet}
\centering
\begin{minted}[fontsize=\small,frame=single,breaklines=true,breakafter=/]{toml}
[workspace]
channels = [
    "https://prefix.dev/pixi-build-backends", 
    "https://prefix.dev/conda-forge", 
    "https://prefix.dev/robostack-staging"]
platforms = ["osx-arm64", "linux-64", "win-64", "linux-aarch64"]
preview = ["pixi-build"]

[tasks]
sim = "ros2 run turtlesim turtlesim_node"
talker = "ros2 run talker_py talker"
navigator = "ros2 run navigator navigator"

[dependencies]
ros-humble-desktop = ">=0.10.0,<0.11"
# Path dependencies
ros-humble-navigator = { path = "src/navigator" }
ros-humble-navigator-py = { path = "src/navigator_py" }
ros-humble-talker-py = { path = "src/talker-py" }
\end{minted}
\caption{Simplified ROS workspace configuration using path dependencies to pixi-build packages. These path dependencies contain a \texttt{pixi.toml} that defines which backend to use. For ROS packages, the backend automatically detects package metadata from \texttt{package.xml} files and handles cross-platform dependency resolution, reducing configuration complexity compared to traditional ROS build workflows.}
\label{fig:ros_pixi_toml}
\end{codesnippet}

The architectural simplicity of pixi-build's backend system is illustrated in Code Snippet~\ref{fig:ros_pixi_toml}, which demonstrates a complete ROS workspace configuration that automatically manages multiple packages with heterogeneous build requirements. This declarative specification replaces the traditional workflow requiring complex colcon or catkin build scripts, multiple dependency specification files, and platform-specific installation procedures. The backend automatically extracts package metadata from each ROS package's \texttt{package.xml} specification, resolves dependencies through RoboStack's conda package mappings, and generates cross-platform conda packages. This approach reduces a traditionally multi-step, error-prone process to a single \texttt{pixi build} command that maintains consistency across development and deployment environments while producing standard \texttt{.conda} packages suitable for direct distribution, eliminating downstream compilation requirements for package consumers.

\subsection*{Reproducible Package Building}
Reproducible package building represents a critical requirement for scientific software distribution, ensuring that packages built at different times, locations, or by different maintainers, produce bit-for-bit identical results. This capability directly supports scientific reproducibility by guaranteeing that package distributions remain consistent regardless of build infrastructure variations.

Both pixi-build and rattler-build implement comprehensive reproducibility measures throughout the build process. File processing follows deterministic ordering to eliminate filesystem-dependent variations. Timestamp information is normalized using the \texttt{SOURCE\_DATE\_EPOCH} standard\footnote{\url{https://reproducible-builds.org/docs/source-date-epoch/}}, ensuring consistent modification times across builds.

Build environment specifications are captured and stored within package metadata, enabling exact environment reconstruction for debugging or verification purposes. Both pixi-build and rattler-build automatically record complete information about how packages were built—including all dependency versions and configuration settings—inside each package as \texttt{info/recipe/rendered\_recipe.yaml}. This means researchers can recreate the exact same build environment years later, which is crucial for long-term scientific reproducibility. Additionally, each package includes cryptographic checksums that verify the package has not been corrupted or altered during distribution, ensuring that downloaded packages match exactly what was originally built, increasing trust in collaborative scientific computing.

Reproducible builds are still an area of active research. A designated repository at \url{https://github.com/prefix-dev/reproducible-builds} tracks the reproducibility status of 100 exemplary packages, of which currently 84, 76 and 34 can be reproduced bit-for-bit on macOS, Linux and Windows respectively (as of September 2025).

\subsection*{Scientific Community Integration and Distribution}
The package creation capabilities of pixi-build and rattler-build integrate with broader scientific computing infrastructure to support diverse research community needs. Custom channel creation enables research groups to distribute specialized packages—including proprietary or lab-specific tools alongside public packages, while maintaining compatibility with standard conda ecosystem tooling. Research groups frequently require domain-specific customization: computer vision researchers need OpenCV builds with specialized hardware acceleration, robotics teams require ROS packages with custom message definitions, perception algorithms demand optimized PCL builds for real-time processing, and autonomous systems need specialized navigation stacks with custom cost functions.

Integration with continuous integration platforms enables automated package building and testing workflows essential for maintaining large package collections. Research groups can establish private or public package channels using standard continuous integration infrastructure, reducing operational overhead while enabling seamless integration of hardware-specific optimizations, such as CUDA-accelerated SLAM libraries, while maintaining packaging standards for community contributions and cross-platform deployment.

Active channels demonstrate the practical application of these capabilities across scientific domains. The rust-forge channel\footnote{\url{https://prefix.dev/channels/rust-forge}} provides Rust language packages across Windows, macOS, and Linux platforms using GitHub Actions integration. Similar approaches support R language packages\footnote{\url{https://prefix.dev/channels/r-forge}} and SLAM tools\footnote{\url{https://anaconda.org/fontan}} (see the case study on VSLAM-LAB in the next Section), illustrating the versatility of pixi-build for diverse research communities.

\section*{Case Studies}

The following case studies demonstrate Pixi's impact through unified package management on robotics research workflows, spanning individual projects to community-wide research infrastructure and industrial deployment. We selected these representative examples from our research collaborations and industry partnerships to illustrate different aspects of Pixi's capabilities across diverse robotics applications. First, VPRTempo illustrates how eliminating platform-specific installation complexity accelerates research collaboration between academic institutions. Second, a MATLAB-based control system shows how Pixi enables cross-platform reproducibility for hybrid MATLAB/C++ research workflows that were previously brittle and OS-specific. Third, VSLAM-LAB demonstrates how unified dependency management enables previously near-impossible systematic comparisons of algorithmic approaches across multiple visual SLAM implementations. Fourth, deploying machine learning models on robotic platforms like the Blue Robotics ROV shows how Pixi can replace complex Docker workflows, significantly lowering barriers for students and researchers to deploy and test models in field conditions. Finally, Smart Robotics' adoption across their 100+ deployed factory robots illustrates how Pixi transforms industrial ROS-based systems, reducing continuous integration times from 120+ minutes to 2-10 minutes while enabling atomic deployments and cross-platform tooling access for non-developer teams. These examples highlight Pixi's role as foundational package-management infrastructure that shifts focus from systems administration to scientific innovation across the entire robotics development lifecycle.

\subsection*{Case Study 1: VPRTempo -- From Complex Installation to Unified Workflow}

Thousands of scientific computing repositories suffer from lengthy, platform-specific installation instructions that create barriers to research collaboration and reproducibility. These instructions typically require users to navigate different commands for various operating systems and hardware configurations, leading to frequent setup failures and support burdens that divert researcher attention from scientific work. VPRTempo~\cite{hines2024vprtempo}, a Spiking Neural Network system for Visual Place Recognition and robotic localization, exemplifies this challenge. We review how Pixi's adoption simplified the installation procedure, led to better reproducibility, and enabled easier collaboration.

\subsubsection*{Traditional Installation Complexity}
Prior to adopting Pixi, VPRTempo required platform-specific installation procedures that created significant barriers to research collaboration and reproducibility. The original documentation required users to navigate four distinct installation pathways:

\begin{enumerate}
\item \textbf{Linux/macOS CPU-only:} \texttt{conda create -n vprtempo -c conda-forge vprtempo}
\item \textbf{Linux CUDA-enabled:} \texttt{conda create -n vprtempo -c conda-forge -c pytorch -c nvidia vprtempo pytorch-cuda cudatoolkit}
\item \textbf{Windows CPU-only:} \texttt{conda create -n vprtempo -c pytorch python pytorch torchvision torchaudio cpuonly prettytable tqdm numpy pandas matplotlib requests}
\item \textbf{Windows CUDA-enabled:} \texttt{conda create -n vprtempo -c pytorch -c nvidia python torchvision torchaudio pytorch-cuda=11.7 cudatoolkit prettytable tqdm numpy pandas matplotlib requests}
\end{enumerate}

Such fragmentation creates several challenges for the research community. Users often encounter environment setup failures due to platform-specific dependency conflicts, particularly with CUDA installations. This documentation burden required maintaining separate installation procedures for each platform, introducing inconsistencies and errors. New collaborators often require substantial technical support to achieve working installations, delaying research contributions.

Additionally, the original workflow required manual dataset management, where users needed to download datasets separately and organize directory structures manually. This process introduced additional opportunities for setup failures and inconsistent experimental configurations across research groups.

\subsubsection*{Pixi-Based Transformation}
The migration to Pixi eliminated the complexity of platform-specific installations through a unified \texttt{pixi.toml} configuration that handles all platform and hardware variations automatically. The new approach demonstrates several key improvements in scientific software distribution:

\textbf{Unified Environment Specification:} A single configuration file specifies dependencies across all supported platforms (linux-64, osx-arm64, win-64) while maintaining platform-specific optimizations through feature systems. The base environment includes essential dependencies with appropriate version constraints: \texttt{python = ">=3.6,<3.13"}, \texttt{pytorch = ">=2.4.0"}, and scientific computing libraries including \texttt{numpy}, \texttt{pandas}, and \texttt{matplotlib}.

\textbf{Hardware-Aware Feature Management:} The CUDA feature system demonstrates sophisticated hardware integration by automatically installing GPU-specific dependencies and CUDA toolkit components only on systems that support them. This approach ensures optimal performance on GPU-enabled systems while maintaining compatibility across diverse hardware configurations.

\textbf{Automated Workflow Integration:} Task definitions replace manual execution procedures with declarative workflow specifications. The \texttt{demo} task automatically downloads pre-trained models and datasets while executing the complete evaluation pipeline via \texttt{pixi run python main.py -{}-PR\_curve -{}-sim\_mat}. Specialized tasks support different research workflows including quantized model evaluation (\texttt{eval\_quant}) and paper replication (\texttt{nordland\_train}, \texttt{oxford\_eval}).

\subsubsection*{Research Impact and Benefits}
The complete research workflow—including installation, dataset download, training, and evaluation—now requires just three commands and completes in under 3 minutes, compared to the previous 30 to 45 minute setup process. The automated dataset download and model retrieval integrated into the demo task eliminated manual setup procedures that previously required detailed documentation and technical support. Documentation maintenance effort decreased from four separate installation procedures to a single workflow description, enabling more focus on research content rather than system administration.

\subsection*{Case Study 2: Reproducibility for academic papers using MATLAB}

Many robotics and control papers implement their proposed algorithms in the proprietary MATLAB language while relying on high-performance C++ libraries that integrate with MATLAB's Foreign Function Interface (FFI), called \texttt{mex}. These hybrid stacks are powerful but often difficult to use across platforms, especially when interfaces depend on system C++ libraries that must be correctly discovered at runtime.

An exemplary case is~\cite{bergonti2022modeling}, which presents the kinematics and control of morphing covers. There, the MATLAB bindings of the C++ library CasADi~\cite{andersson2019casadi} are used. The addition of Pixi to this repository drastically simplified setup and reproducibility across Windows, macOS, and Linux.\footnote{\href{https://github.com/ami-iit/paper_bergonti_2022_tro_kinematics-control-morphingcovers/pull/11}{https://github.com/ami-iit/paper\_bergonti\_2022\_tro\_kinematics-control-morphingcovers/pull/11}}

Before integrating Pixi, users needed to manually manage their MATLAB libraries. The steps required to install MATLAB libraries varied between libraries, often requiring expert intervention. Furthermore, manually installed libraries were difficult to version-lock.

\subsubsection*{Pixi Integration}

The first step of migrating the repository to Pixi was to ensure that the required MATLAB and C++ libraries were available for use with pixi. The CasADi~\cite{andersson2019casadi} C++ library was already available in conda-forge. The other libraries required were \texttt{casadi-matlab-bindings}, the MATLAB bindings for the CasADi C++ library, and \texttt{mystica}, a pure MATLAB library developed by the paper's authors.

The conda packages for the \texttt{casadi-matlab-bindings} were automatically created using GitHub Actions, so they can be easily and automatically updated for new versions of CasADi. To ensure that the MATLAB bindings contained in the package could be found by MATLAB, the package's activation script appended the location of the library to the \texttt{MATLABPATH} environment variable.

As the \texttt{mystica} library is a pure MATLAB library, it was downloaded and installed on the system using a Pixi task, and its location was appended to the \texttt{MATLABPATH} environment variable in the \texttt{pixi.toml} activation table.

This setup resulted in the following advantages:
\begin{itemize}
  \item \textbf{Unified dependency specification:} Pixi installs the CasADi C++ library and headers from conda-forge, pinned via the lockfile, ensuring consistent builds across collaborators and continuous integration.
  \item \textbf{MATLAB-aware activation:} Platform-specific activation sets search paths so MATLAB can discover CasADi's shared libraries without modifying the MATLAB installation.
  \item \textbf{Reproducible tasks:} Simulation experiments of the paper can be reproduced by simply running a single command, \texttt{pixi run sim1}. If MATLAB is not available on the PATH, the tasks fail with a clear error message.
\end{itemize}

This design respects licensing constraints (Pixi does not distribute or install MATLAB) while providing a reproducible environment for all the libraries that can be legally distributed.

\subsubsection*{Outcomes}

The MATLAB + C++ integration, previously a source of fragile, OS-specific instructions, becomes a one-command experience:\ \texttt{pixi run run\_paper}. Benefits observed by maintainers and users include:
\begin{itemize}
  \item \textbf{Onboarding time reduced} from hours to minutes; no manual compiler or path setup.
  \item \textbf{Cross-platform consistency} via a single lockfile that pins CasADi and its dependencies.
  \item \textbf{Headless reproducibility} in continuous integration using \texttt{matlab-batch}\footnote{\url{https://www.mathworks.com/help/parallel-computing/batch.html}} and Pixi environments.
  \item \textbf{Lower maintenance burden} as documentation converges to a tiny set of Pixi tasks rather than per-operating-system guides.
\end{itemize}
This case demonstrates that Pixi is not limited to Python-centric workflows: it provides reliable C/C++ userlands that proprietary tools like MATLAB can consume, turning historically brittle MATLAB+MEX stacks into reproducible, shareable research artifacts.

\subsection*{Case Study 3: VSLAM-LAB -- Enabling Unified SLAM System Benchmarking}

Robotics research across multiple domains suffers from algorithmic fragmentation, where different implementations within the same field require distinct toolchains, dependencies, and evaluation methodologies. Researchers attempting comparative studies must navigate incompatible build systems, conflicting dependencies, and inconsistent evaluation protocols, preventing systematic benchmarking and hindering progress in understanding relative algorithm performance.

Visual Simultaneous Localization and Mapping (VSLAM) exemplifies this broader challenge. VSLAM-LAB~\cite{FontanIROS2025} demonstrates how unified package management can transform fragmented research domains by creating comprehensive benchmarking infrastructure that was previously infeasible to develop. VSLAM-LAB addresses this fundamental challenge by creating the first unified framework for VSLAM benchmarking, integrating both state-of-the-art systems (MASt3R-SLAM, DVPO, Mono-GS, DROID-SLAM) and established baselines (ORB-SLAM2, DSO) alongside Structure-from-Motion pipelines (COLMAP, GLOMAP, DUSt3R, SPANN3R) within reproducible environments, enabled by Pixi.

\subsubsection*{Technical Integration Challenge}
The development of VSLAM-LAB required substantial engineering effort to unify disparate SLAM systems that were originally designed with incompatible dependency requirements and build procedures. Each SLAM method typically demanded specific versions of computer vision libraries, different Python environments, and platform-specific compilation procedures. Traditional approaches would require researchers to maintain separate environments for each algorithm, making comparative evaluation prohibitively complex.

VSLAM-LAB embraces modularity by preserving one environment per implemented baseline. Pixi's seamless environment switching allows multiple baselines with incompatible dependencies to coexist within the same benchmarking framework. The framework's \texttt{pixi.toml} configuration integrates diverse requirements including OpenCV builds with specific feature sets, PyTorch installations with CUDA compatibility, ROS components for sensor data handling, and specialized mathematical libraries required by different algorithms. This design avoids the need to standardize dependencies across baselines, significantly reducing integration overhead while preserving reproducibility and fidelity to original implementations.

\subsubsection*{Unified Evaluation Workflow for Seamless Reproducibility}
The resulting framework enables unprecedented ease and reproducibility in the evaluation of Visual SLAM systems through a unified command interface. By leveraging Pixi's task system and multi-environment management, users can execute any supported baseline on any dataset sequence using a simple, consistent syntax: \texttt{pixi run demo <baseline> <dataset> <sequence>}

This interface abstracts the complexity of individual system requirements, enabling comparative evaluations that were previously impractical. Examples include:
\begin{itemize}
\item \texttt{pixi run demo mast3rslam eth table\_3}
\item \texttt{pixi run demo droidslam euroc MH\_01\_easy} 
\item \texttt{pixi run demo orbslam2 rgbdtum rgbd\_dataset\_freiburg1\_xyz}
\end{itemize}

VSLAM-LAB automates the entire benchmarking pipeline, including dataset downloading, preprocessing, dependency setup, algorithm execution, and result evaluation. Researchers define their experiments in a simple YAML file; the system ensures consistent execution across different machines and users. The standardized evaluation methodology promotes reproducible research by ensuring consistent experimental conditions across studies. Researchers can now focus on algorithmic development rather than system integration, accelerating progress toward robust, real-world SLAM solutions.

\subsubsection*{Research Community Impact}
VSLAM-LAB enables systematic comparative studies that were previously infeasible due to integration complexity. The unified framework allows researchers to evaluate algorithmic innovations across multiple baselines and datasets without the traditional barriers of environment management and system integration.

Thanks to Pixi's robust dependency handling and lockfile system, SLAM baselines can now be executed off-the-shelf with any dataset, which empowers both rapid experimentation and large-scale benchmarking. For example, a researcher evaluating SLAM solutions for a new application can define a dataset once and run it across all available baselines without additional setup.

To illustrate the typical overhead, integrating a single SLAM baseline with a dataset from scratch could take up to 8 hours. As a result, running 15 baselines on a custom dataset might require around 120 hours, and scaling this to 20 datasets could demand up to 2,400 hours. VSLAM-LAB reduces this overhead to a constant, limited only to defining the experiment in a YAML file—regardless of the number of baselines or datasets. Furthermore, Pixi's no-root-privilege design enables seamless deployment on high-performance computing clusters, while the framework's architecture allows inference tasks to be easily parallelized across multiple nodes or GPUs. This substantially reduces setup time and effort, enabling scalable benchmarking with minimal overhead and efficient resource utilization.

VSLAM-LAB's extensibility supports community contributions, where new algorithms can be integrated into the existing evaluation infrastructure through standardized interfaces. This approach establishes a sustainable platform for ongoing SLAM research, benefiting the entire community.

\subsection*{Case Study 4: Deploying Machine Learning on Robots -- Replacing Docker with Pixi}

\begin{figure}
    \centering
    \includegraphics[width=0.6\linewidth,clip,trim=3cm 2cm 3cm 5cm]{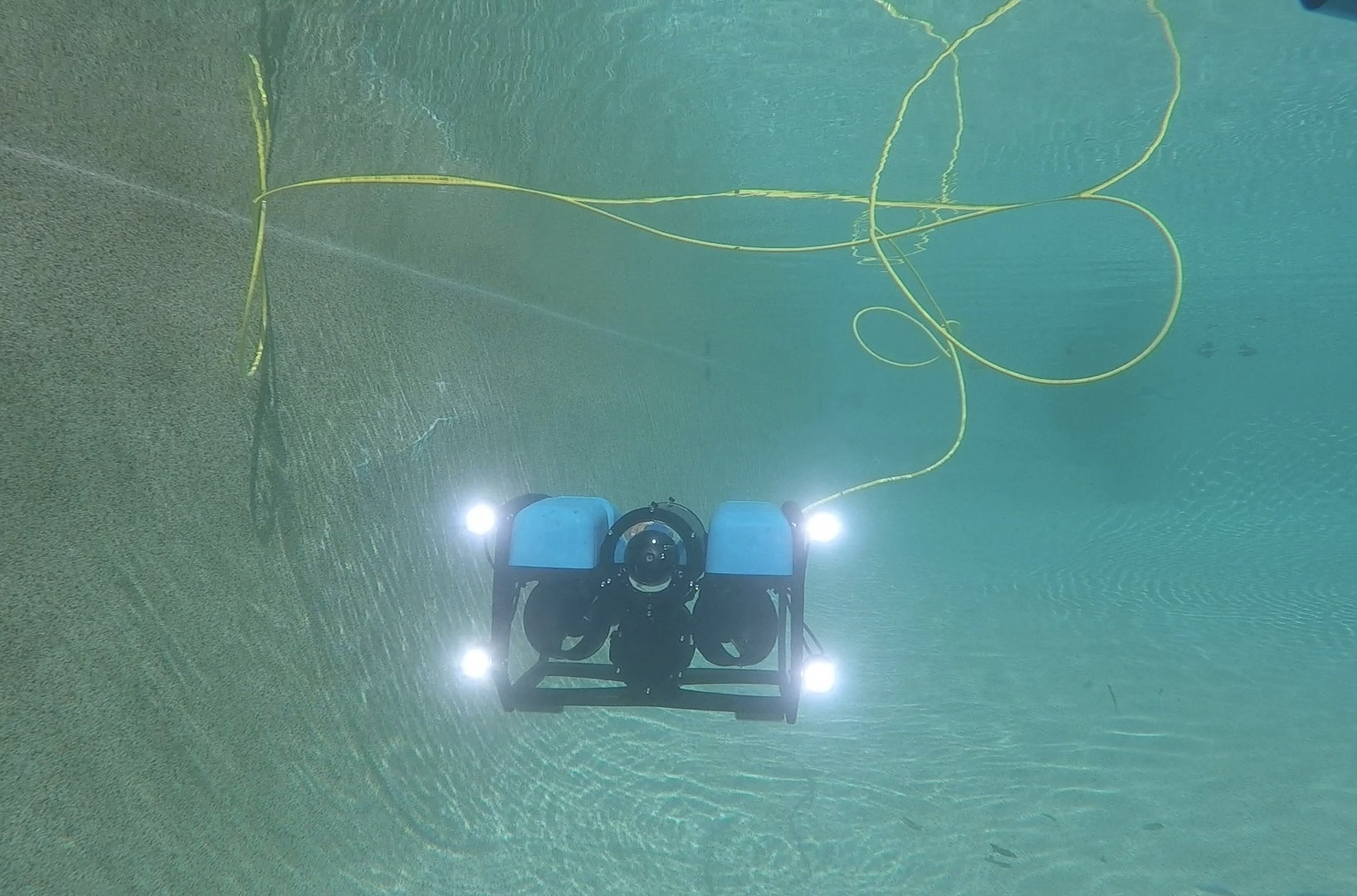}
    \caption{Blue Robotics Remotely Operated Vehicle}
    \label{fig:bluerov}
\end{figure}

Deploying machine learning (ML) models on robotic systems like the Blue Robotics ROV (Fig.~\ref{fig:bluerov}) traditionally involves complex Docker workflows and deep familiarity with the Robot Operating System (ROS) and related tools. While powerful, Docker-based environments can be difficult to configure, particularly for students or researchers without extensive DevOps experience. This setup complexity often becomes a barrier to adoption, slowing down model iteration and real-time testing in the field. Shifting to Pixi eliminates the overhead of containerization: Pixi allows for streamlined, Python-native environments that launch quickly and integrate seamlessly with ROS and ML toolkits. This enables students to run and evaluate models on the ROV in real-time, without needing to navigate Dockerfile creation, container management, or ROS setup intricacies.

\subsubsection*{Original Docker Workspace}
The original system for deploying machine learning models on the Blue Robotics ROV was built around a Docker-based workspace (Code Snippets~\ref{fig:dockerfile} and \ref{fig:devcontainer}). While this approach offered environment reproducibility and compatibility with ROS and CUDA, it introduced significant complexity. Users needed to install and configure Docker, manage development containers, install CUDA drivers, and navigate the structure of ROS packages within a fixed directory layout. This steep learning curve made the system difficult to use, especially for students or researchers unfamiliar with Docker, ROS, or containerized development. Installing the required tooling could take hours, with common pitfalls around permissions, image compatibility, and hardware acceleration.

\begin{codesnippet}
\centering
\begin{minipage}{0.95\linewidth}
\begin{minted}[fontsize=\small,frame=single,breaklines]{dockerfile}
FROM osrf/ros:jazzy-desktop-full

ARG USERNAME=ros
ARG USER_UID=1000
ARG USER_GID=$USER_UID

# Delete "ubuntu" user if it exists, then create a new user
RUN if getent passwd ubuntu > /dev/null 2>&1; then \
    userdel -r ubuntu && echo "Deleted existing ubuntu user"; fi && \
    groupadd --gid $USER_GID $USERNAME && \
    useradd -s /bin/bash --uid $USER_UID --gid $USER_GID -m $USERNAME && \
    echo "Created new user $USERNAME"

# Add sudo support
RUN apt-get update && apt-get install -y sudo \
 && echo $USERNAME ALL=\(root\) NOPASSWD:ALL > /etc/sudoers.d/$USERNAME \
 && chmod 0440 /etc/sudoers.d/$USERNAME && rm -rf /var/lib/apt/lists/*

# GStreamer plugins for video streaming
RUN apt-get update && apt-get install -y \
 python3-gst-1.0 gstreamer1.0-plugins-good \
 gstreamer1.0-plugins-bad gstreamer1.0-libav

USER $USERNAME
ENV WORKSPACE_HOME=/docker_ws

# Add bash aliases for ROS
RUN echo " Adding aliases to bashrc" \
 && echo "alias sros='source /opt/ros/jazzy/setup.bash'" >> /home/$USERNAME/.bashrc \
 && echo "alias sws='sros && source /docker_ws/install/setup.bash'" >> /home/$USERNAME/.bashrc \
 && echo "if [ -f /opt/ros/jazzy/setup.bash ]; then source /opt/ros/jazzy/setup.bash; fi" >> /home/$USERNAME/.bashrc \
 && echo "if [ -f /docker_ws/install/setup.bash ]; then source /docker_ws/install/setup.bash; fi" >> /home/$USERNAME/.bashrc
\end{minted}
\end{minipage}
\caption{Dockerfile for the ROS 2 Jazzy environment used in development. It installs GStreamer plugins for video streaming, sets up a custom non-root user, and adds useful ROS sourcing aliases. This setup supports GPU-based perception models and live ROV telemetry.}
\label{fig:dockerfile}
\end{codesnippet}

\begin{codesnippet}
\centering
\begin{minipage}{0.95\linewidth}
\begin{minted}[fontsize=\small,frame=single,breaklines=true,breakafter=/]{json}
{
  "name": "docker_ws",
  "build": {
    "dockerfile": "Dockerfile",
    "context": ".."
  },
  "mounts": ["type=bind,source=/etc/localtime,destination=/etc/localtime,readonly",
    "type=bind,source=/tmp/.X11-unix,destination=/tmp/.X11-unix",
    "type=bind,source=
        ${localEnv:XAUTHORITY},
        destination=/home/vscode/.Xauthority"
  ],
  "containerEnv": {
    "DISPLAY": "${localEnv:DISPLAY}",
    "XAUTHORITY": "/home/vscode/.XAuthority",
    "NVIDIA_DRIVER_CAPABILITIES": "compute,utility,graphics",
    "NVIDIA_VISIBLE_DEVICES": "all"
  },
  "runArgs": [
    "--device", "/dev/dri:/dev/dri", "--net", "host", "--ipc", "host",
    "--pid", "host", "--privileged", "--runtime", "nvidia", "--name", "docker-dev"
  ],
  "postCreateCommand": "sudo chown -R vscode /ccache",
  "remoteUser": "ros",
  "workspaceFolder": "/docker_ws",
  "workspaceMount": "source=${localWorkspaceFolder},\
        target=/docker_ws,type=bind"
}
\end{minted}
\end{minipage}
\caption{\texttt{devcontainer.json} for ROS development in VS Code. This config enables NVIDIA GPU passthrough, X11 GUI support, and persistent user environment with Python and C++ tooling. Used alongside the Dockerfile in Code Snippet~\ref{fig:dockerfile}.}
\label{fig:devcontainer}
\end{codesnippet}

This setup, while powerful for advanced users, created unnecessary barriers for early-stage students and non-specialist collaborators. Even minor configuration issues could derail development and deployment workflows.

\subsubsection*{Simplified Workflow Without Containerization}
The Pixi-based setup (Code Snippet~\ref{fig:pixi_config_clean}) improves on the previous Docker-based workflow by eliminating the need for users to install and configure Docker, VSCode extensions, or CUDA user-space libraries. Instead of relying on containers and complex system-level permissions, Pixi creates lightweight, reproducible environments that work across different operating systems (in this example, Linux and macOS) without requiring administrative access. All dependencies, including ROS, machine learning libraries, and GStreamer plugins, are managed declaratively in a single \texttt{pixi.toml} file. This simplifies onboarding, streamlines debugging, and empowers students and researchers to build, run, and test their models with a single command, without needing prior experience with Docker or ROS setup.

\begin{codesnippet}
\centering
\begin{minipage}{0.95\linewidth}
\begin{minted}[fontsize=\scriptsize,frame=single,breaklines=true,breakafter=/]{toml}
[project]
name = "MLonBlueRoboticsROV"
version = "0.1.0"
description = "Pixi-based ROS ML workspace"
authors = ["Author Name <contact@example.com>"]
channels = ["conda-forge",
    "https://fast.prefix.dev/conda-forge",
    "https://prefix.dev/robostack-jazzy"]
platforms = ["linux-64", "osx-arm64"]
preview = ["pixi-build"]

[system-requirements] 
libc = { family = "glibc", version = "2.22" }
cuda = "12"

[dependencies]
python = "*"
compilers = "*"
cmake = "*"
pkg-config = "*"
make = "*"
ninja = "*"
ros-jazzy-desktop = "*"
ros-jazzy-cv-bridge = "*"
colcon-common-extensions = "*"
rosdep = "*"
...

[target.linux.dependencies]
pytorch-gpu = "*"
gstreamer = ">=1.24.6,<2"
...

[target.osx-arm64.dependencies]
ipykernel = ">=6.29.5,<7"
pygobject = "*"
gstreamer = "*"
...

[activation]
scripts = ["install/setup.bash"]

[activation.env]
ROS_DOMAIN_ID = "23"
YOLO_VERBOSE = "False"
RMW_IMPLEMENTATION = "rmw_cyclonedds_cpp"

[tasks]
build = { cmd = "colcon build --merge-install --symlink-install --cmake-args '-DCMAKE_BUILD_TYPE=RelWithDebInfo' -Wall -Wextra -Wpedantic" }

rosbag_record = {cmd = "./scripts/record_rosbag.sh"}
streamer = {cmd = "ros2 run video_stream_processor nros_streamer"}

yolo_ros = {cmd = "ros2 launch example_pkg yolo_ros_example.xml"}
yolo_nros = {cmd = "ros2 run example_pkg online_yolo_example --save_img True"}
rviz2_yolo = {cmd = "rviz2 -d config/rviz_yolo_ros.rviz"}
\end{minted}
\end{minipage}
\caption{\texttt{pixi.toml} configuration for the simplified ROS and ML workspace. The Pixi environment is portable, cross-platform, and removes barriers such as Docker or system dependencies. All dependencies, GPU tooling, and task commands are unified under a single reproducible environment.}
\label{fig:pixi_config_clean}
\end{codesnippet}

\subsubsection*{Research Community Impact}
For the field robotics research community, adopting Pixi provides a more inclusive and agile approach to robotics experimentation. By abstracting away the complexity of Docker and ROS setup, Pixi makes it easier for interdisciplinary researchers, including those from ecology, environmental science, and education, to engage with robotic vision workflows. This fosters broader collaboration and faster prototyping of machine learning models for real-world deployments. Ultimately, Pixi accelerates research cycles, improves reproducibility, and broadens access to robotic platforms for real-time, in-situ environmental monitoring and AI-driven decision-making.

\subsection*{Case Study 5: Smart Robotics}

\begin{figure}
     \centering
     \begin{subfigure}[b]{0.3\textwidth}
         \centering
         \includegraphics[width=\textwidth]{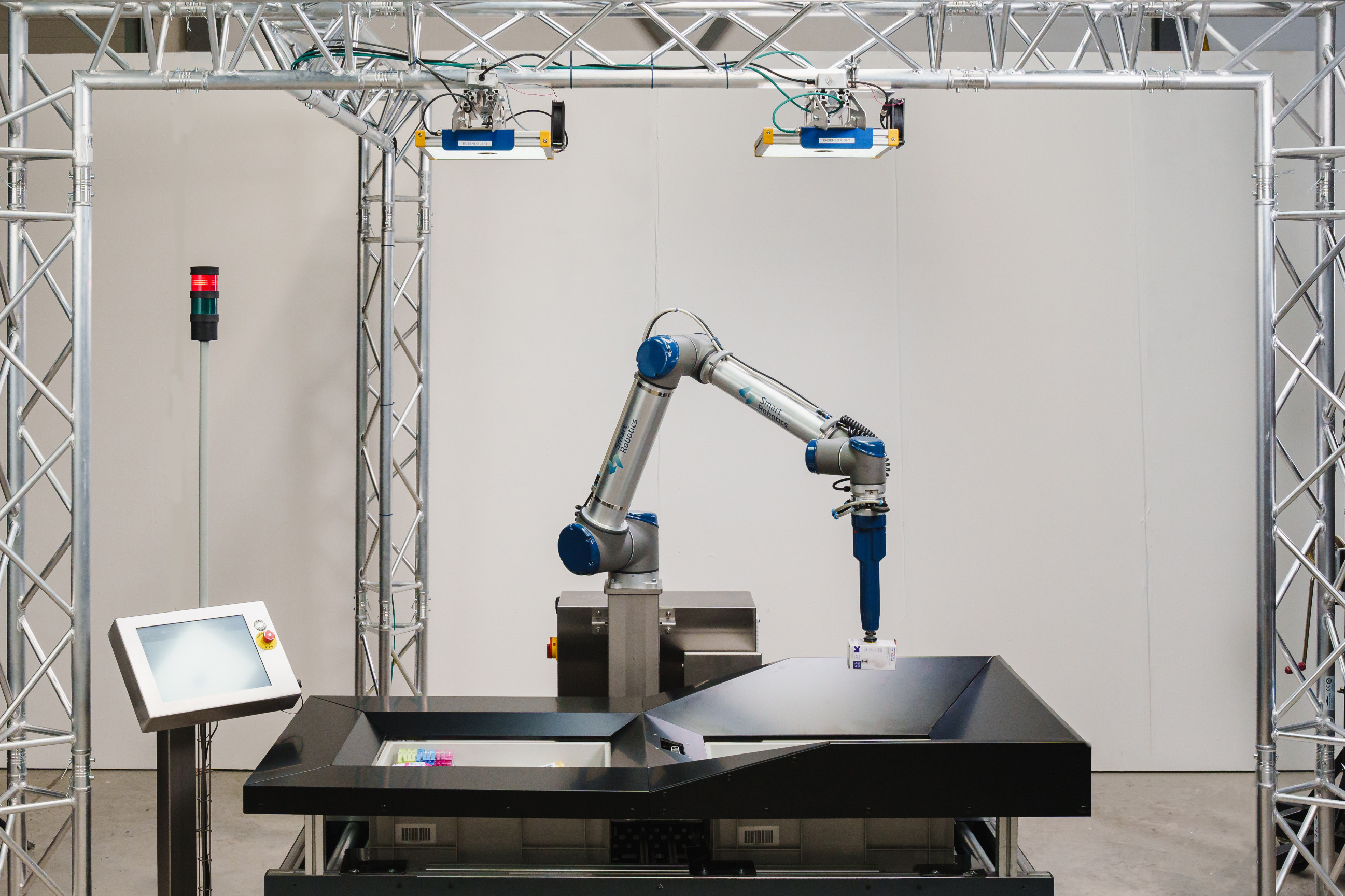}
         \caption{Cobot Item Picker}
         \label{fig:item picker}
     \end{subfigure}
     \hfill
     \begin{subfigure}[b]{0.3\textwidth}
         \centering
         \includegraphics[width=\textwidth]{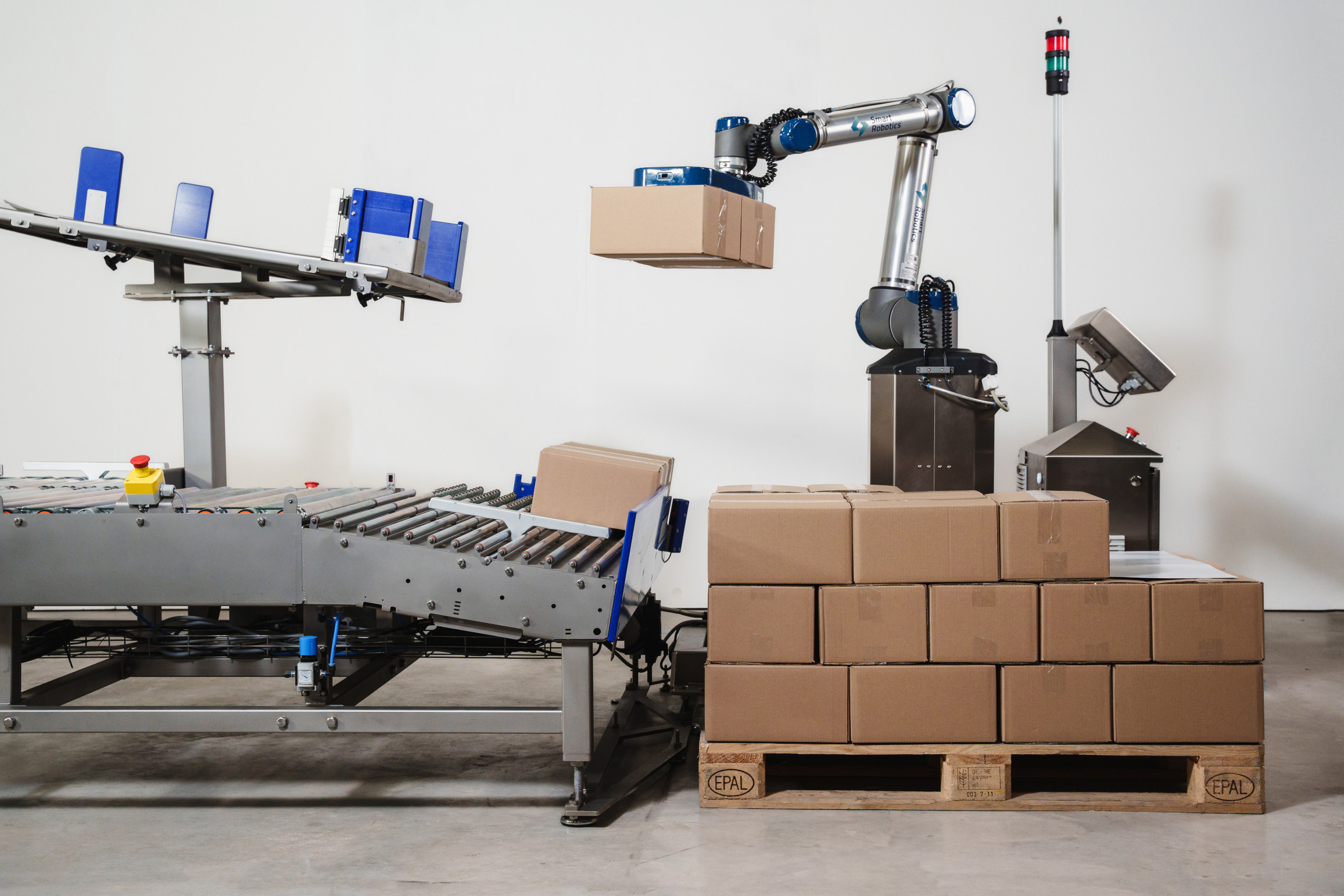}
         \caption{Cobot Palletizer}
         \label{fig:palletizer}
     \end{subfigure}
     \hfill
     \begin{subfigure}[b]{0.3\textwidth}
         \centering
         \includegraphics[width=\textwidth]{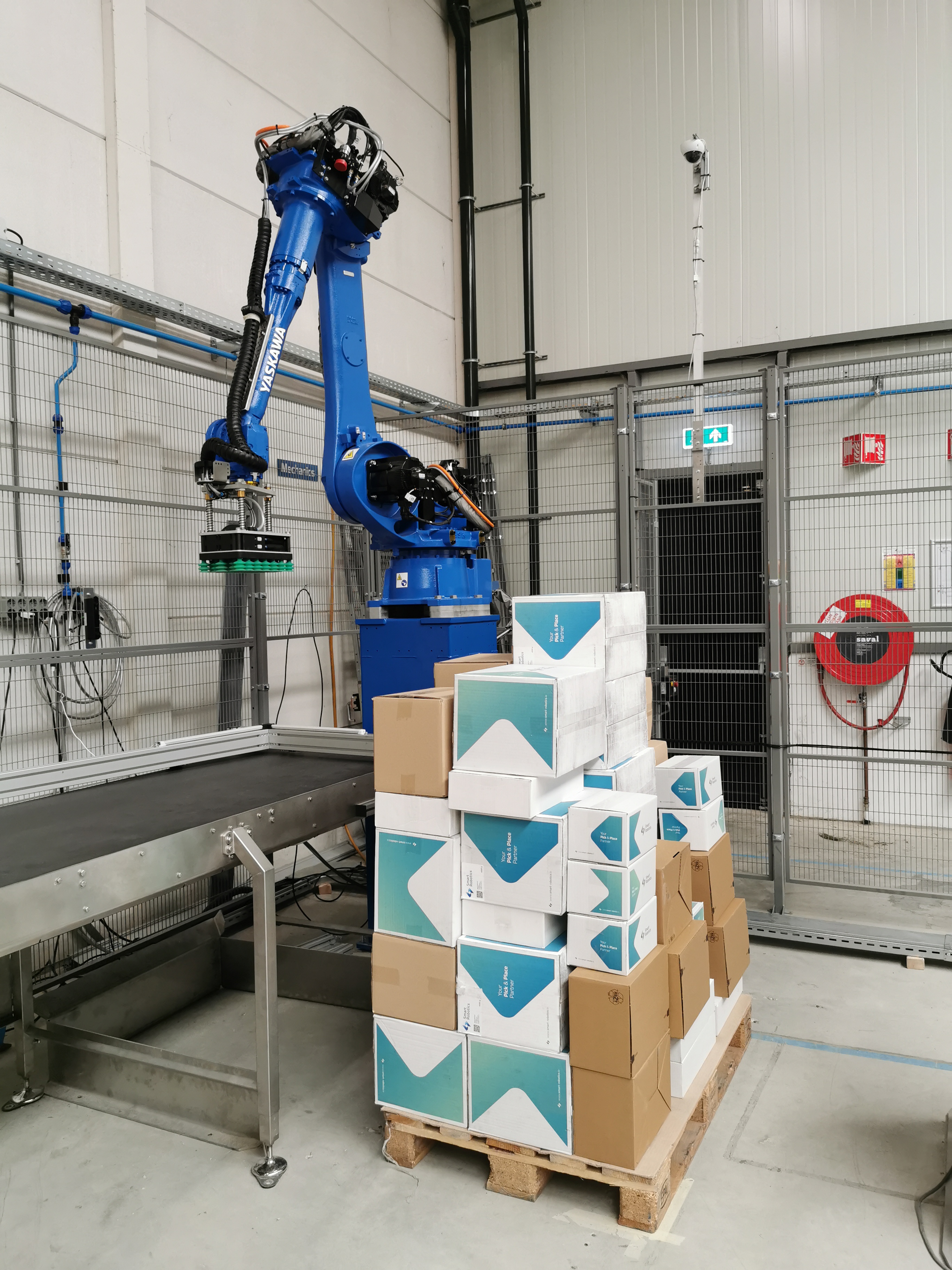}
         \caption{Industrial Depalletizer}
         \label{fig:depalletizer}
     \end{subfigure}
        \caption{Example of robots from Smart Robotics powered by \texttt{pixi}.}
        \label{fig:robots powered by pixi}
\end{figure}

Smart Robotics is a robotics company located in the Netherlands that specializes in automated pick-and-place solutions. They have more than 100 robots deployed in factories and logistics centers worldwide. Since 2015, Smart Robotics has relied on ROS to power its robot systems, which now operate primarily on Universal Robots cobots and Yaskawa Motoman robots (see examples in Fig.~\ref{fig:robots powered by pixi}).

Smart Robotics has invested in a large, custom-built robotics software stack, which includes collision-aware motion planning and execution, world modeling, vision, and decision making. Around 20 developers contribute to this stack across multiple projects, often needing to switch quickly between projects and ROS versions, for example when supporting systems in the field or developing different products.

\subsubsection*{Pre pixi workflow}
\begin{codesnippet}
\centering
\begin{minipage}{0.95\linewidth}
\begin{minted}[fontsize=\small,frame=single,breaklines,breakafter=/]{yaml}
packages:
  pybind11:
  - bash-src:
      check_presence_script: |
        #!/bin/bash
        source /opt/sr/rosdep_functions.bash

        # Check if it is installed and has the right version
        sr-rosdep-check-md5sum /usr/local/include/pybind11/detail/common.h 750d572 || exit
      install_script: |
        #!/bin/bash
        rm /tmp/pybind11 -rf
        git clone https://github.com/pybind/pybind11 /tmp/pybind11 v2.2.2
        mkdir /tmp/pybind11/build
        cd /tmp/pybind11/build
        cmake .. -DPYBIND11_TEST=false
        make && sudo make install
        rm /tmp/pybind11 -rf
    version: 2.2.2
    depends_on:
      cmake: {}
    type: bash-src
\end{minted}
\end{minipage}
\caption{Example custom \texttt{rosdep} configuration for installing pybind.}
\label{fig:rosdep-bash-src}
\end{codesnippet}

Originally, Smart Robotics organized its code in a handful of large mono-repositories. 
For development this approach meant that the large repositories with deep dependency trees needed to be installed globally, which often led to issues and broken packages when switching between major versions. 
Deployments required cloning source code onto production machines, installing external dependencies globally via rosdep (using apt, pip, and custom rdmanifest rules, see an example in Code Snippet~\ref{fig:rosdep-bash-src}), and compiling directly on-site. After installation, the source code was discarded for proprietary reasons. Upgrades required a global reinstall of dependencies, which was fragile and difficult to roll back if something failed. Continuous integration suffered as well: every continuous integration run required rebuilding the entire mono-repository, resulting in lengthy continuous integration cycles.

This approach made it nearly impossible to:
\begin{itemize}
  \item Maintain reproducible setups across Linux distributions, Ubuntu versions, or different ROS releases.
  \item Enable developers to switch projects without breaking existing environments.
  \item Support non-Linux users (e.g.~sales and service teams on Windows) with tooling which runs natively on their machines.
\end{itemize}

\subsubsection*{Migration to Pixi}
To overcome these challenges, Smart Robotics has split its mono-repositories in single packages, and migrated the resulting 120+ packages into the conda ecosystem. The team moved away from global installations and embraced fully user-space environments with lockfiles, enabling truly reproducible deployments across machines. In the beginning, this was done using \texttt{micromamba} and \texttt{boa}\footnote{Boa was a predecessor to rattler-build for creating conda packages.}, and more recently Smart Robotics has adopted \texttt{pixi} and \texttt{rattler-build} as backends for the conda based installation.  

To build and publish a package, \texttt{package.xml} is built with \texttt{pixi-build}, and then published to a self-hosted server. Similarly, third-party dependencies not yet available on conda-forge, a custom \texttt{recipe.yaml} (see an example in Code Snippet~\ref{fig:zenoh_recipe}) is created and the resulting package is published.

\subsubsection*{Outcomes}
As a result, deployments are simplified: a \texttt{package.xml} describing dependencies is automatically built and conda packages created. The resulting packages and their dependencies can be easily installed in user space; no source code or build steps are required on production machines. Upgrades are handled by creating new workspaces that can be switched atomically, with immediate rollback possible if an issue arises.

For the development team, development on multiple projects, each living in its own workspace greatly improved the workflow. The change from \texttt{micromamba} to \texttt{pixi} decreased the solving time needed for complex cases from 80 seconds down to 7 seconds on the first run and to below one second for consecutive solving when a cache is available (see also Fig.~\ref{fig:package_manager_benchmark}). 

Furthermore, continuous integration runs are significantly faster, since only changed packages are rebuilt rather than entire monolithic repositories. The speedup varies from previously 120+ minutes down to 2-3 minutes for the smaller packages and up to 10 minutes for larger test-suites, resulting in far fewer resources needed for testing and shorter wait times during development.

For the wider organization, Pixi's cross-platform reproducibility empowers sales and service departments to use parts of the same robotics stack and tooling on their Windows laptops as the developers use on Linux.

In short, Smart Robotics' migration to conda and Pixi transformed both development and deployment workflows. It brought reproducibility, flexibility, and robustness to a large-scale ROS-based system in production, while also simplifying upgrades, reducing continuous integration overhead, and extending access to non-developer teams.

\section*{Discussion}

These case studies represent just five examples of the 5,300 projects that have adopted Pixi, demonstrating the transformative potential of unified package management across scientific computing. VPRTempo exemplifies how individual research projects benefit from simplified workflows (from 30-45 minutes to just 3 minutes, eliminating setup failures), while the MATLAB/CasADi morphing covers study shows how Pixi transforms historically brittle hybrid language workflows into reproducible, cross-platform research artifacts. VSLAM-LAB illustrates the broader possibility of creating unified research infrastructure that integrates incompatible systems for systematic benchmarking, while the Blue Robotics ROV case ddemonstrates how Pixi streamlines deployment of machine learning models on robotic platforms, making it accessible to beginners and achievable without complex Docker containerization. Finally, Smart Robotics' industrial deployment across 100+ factory robots shows how Pixi scales from individual research projects to production systems, reducing CI times from 120+ minutes to 2-10 minutes while enabling atomic deployments and cross-platform tooling access. %

These advances do not come without challenges. Integrating diverse software ecosystems often requires patches, i.e.~small source code modifications that restore compatibility between older packages and newer compilers or library versions. Legacy scientific software often requires modifications to support modern build systems or updated dependency specifications. However, the conda-forge community has demonstrated a remarkable willingness to address these integration challenges collaboratively. The ecosystem's maintainer network actively supports package updates, compatibility fixes, and integration assistance for research software.

Recent developments in the conda-forge ecosystem have significantly enhanced the foundation upon which Pixi operates. The close collaboration between conda-forge and NVIDIA has brought reliable CUDA 12 support directly to the conda-forge channel, eliminating the need for separate NVIDIA or PyTorch channels that previously complicated dependency management. This integration ensures that GPU-accelerated workflows can be managed through unified channels with consistent dependency resolution.

The broader implications extend beyond technical improvements to fundamental changes in scientific collaboration patterns. Pixi's unified approach addresses the core reproducibility challenges that have hindered scientific progress: inconsistent environments, dependency conflicts, and platform-specific variations. By providing deterministic environment specifications verified cryptographically, the framework enables the bit-for-bit reproducibility required for rigorous scientific validation.

Perhaps most significantly, Pixi enables a shift toward community-driven research infrastructure, where complex integration work is shared and reused across research groups. The success of projects like VSLAM-LAB demonstrates how unified package management can support collaborative development of benchmarking platforms, evaluation frameworks, and shared research tools that benefit entire scientific communities. This model suggests a future where research software development becomes increasingly collaborative and cumulative, rather than fragmented and duplicative.

Despite its rapid adoption, Pixi is not a universal solution. Large technology organizations often operate bespoke internal build and deployment ecosystems, where integration with existing security policies, licensing constraints, or custom hardware toolchains can pose adoption hurdles. Similarly, projects requiring highly specialized real-time operating systems or proprietary compiler environments may not yet integrate smoothly within the current conda-forge and Pixi build infrastructure. As Pixi continues to mature, extending support to these domains will require sustained community effort, institutional engagement and further alignment with industry standards.

The challenges Pixi addresses align with experiences documented in successful long-term scientific software communities. The RosettaCommons, which has grown to over 3 million lines of code across 71 institutions, demonstrates that sustainable collaborative development requires not only technical infrastructure but also shared community standards and coordinated dependency management~\cite{koehler2020better}. Their experience developing version control, testing frameworks, and collaborative workflows over two decades validates the importance of comprehensive package management solutions that reduce technical barriers to scientific collaboration.

\section*{Conclusion}

Pixi has been designed from first principles to address the reproducibility and collaboration challenges that have historically hindered scientific computing and robotics research. Built upon modern dependency resolution algorithms, unified ecosystem integration, and comprehensive environment specification, Pixi provides a robust framework for scientific software management that supports the complex, multi-language requirements of robotics and AI research. The framework continues to accelerate the transition from laboratory prototypes to reproducible, community-accessible research tools that drive scientific progress.

The case studies presented demonstrate how Pixi serves as an enabler, equalizer, and accelerator for scientific research. The standardization of scientific software management around unified principles, creating opportunities for new forms of collaborative research, faster algorithm development, and more effective knowledge transfer across the scientific community. As Pixi continues to mature and expand its capabilities, this trend toward unified and reproducible scientific computing infrastructure is likely to accelerate, fundamentally transforming how researchers develop, share, and build upon computational tools in robotics and beyond.

Looking ahead, Pixi represents a foundational shift toward treating dependency management as core research infrastructure. As robotic systems become increasingly complex and interdisciplinary, unified package management will likely become as essential to research as version control is today. The success of projects like VSLAM-LAB suggests a future where research communities can more rapidly build upon one another’s work, accelerating the transition from laboratory prototypes to deployed systems that benefit society.

\bibliographystyle{sciencemag}
\bibliography{scibib}

\section*{Acknowledgments}
T.F., A.F., A.D.H., M.M., and S.R.~acknowledge continued support from the Queensland University of Technology (QUT) through the Centre for Robotics.
\paragraph*{Funding} This work received funding from an ARC Laureate Fellowship FL210100156 to M.M.~and an ARC Discovery Early Career Researcher Award DE240100149 to T.F.
\ifsciencestyle
\paragraph*{Author contributions} 
T.F.~conceived the study, wrote the main manuscript, and oversaw the research design and execution. W.V., B.Z., R.A., and T.d.J.~designed and implemented Pixi, contributed foundational technical content, and provided critical technical insights throughout manuscript development. A.F., A.H., D.C., S.T., and S.R.~contributed case study analyses and provided domain expertise in robotics applications. T.F., A.F., A.H., M.M., S.T., D.C., and S.R.~critically revised the manuscript and provided substantial feedback on scientific content and presentation. M.M.~and S.R.~provided strategic research guidance and domain expertise in robotics and AI. M.M.~and~S.R. supervised the overall research direction, provided senior oversight of the project conception and execution, and guided the manuscript's strategic positioning and scientific rigor.
\paragraph*{Competing interests} W.V., B.Z., R.A., and T.d.J.~are employees of prefix.dev, the company that develops and maintains the open-source Pixi package manager. Prefix.dev may benefit commercially from increased adoption of Pixi through related consulting services and enterprise support. D.C.~is an employee of Smart Robotics, which uses Pixi in their commercial robotics systems and may benefit from its continued development and adoption. All other authors declare no competing interests.
\paragraph*{Data and materials availability} All data and code used in this study is available at \url{https://github.com/prefix-dev}.
\fi

\ifsciencestyle
\subsection*{Supplementary materials}
The supplementary materials provide detailed technical support for the manuscript’s claims about Pixi’s innovations in package management. They include an analysis of the limitations of existing SAT-based, backtracking, and heuristic dependency resolvers, together with a detailed exposition of Pixi’s advanced dependency-resolution architecture and performance gains.
\fi

\ifsciencestyle
\newpage
\fi

\renewcommand{\thefigure}{S\arabic{figure}}
\renewcommand{\thetable}{S\arabic{table}}
\renewcommand{\theequation}{S\arabic{equation}}
\renewcommand{\thepage}{S\arabic{page}}
\setcounter{figure}{0}
\setcounter{table}{0}
\setcounter{equation}{0}
\setcounter{page}{1} %

\begin{center}
\section*{Supplementary Materials for\\ \scititle}
\def\and{, }
\sciauthornames\\
\small{\sciauthorQUT}\\
\small{\sciauthorPrefix}\\
\small{\sciauthorIIT}\\
\small{\sciauthorSmartRobotics}\\
\small{$^\ast$Corresponding author. Email: tobias.fischer@qut.edu.au}
\end{center}

\subsubsection*{This PDF file includes:}
\begin{itemize}
\item \textbf{Supplementary Figure S1}: Package manager performance comparison across different environments. Error bars show standard deviation. The microenv installs \texttt{bat}, a \texttt{cat} clone with syntax highlighting and Git integration. The dummyenv installs various visualization libraries, i.e.~\texttt{scipy}, \texttt{polars}, \texttt{ipywidgets}, \texttt{seaborn}, \texttt{matplotlib}, \texttt{plotly}, and \texttt{pandoc} (via \texttt{pip}). The stressenv installs \texttt{rust}, \texttt{jupyterlab} and \texttt{pytorch}.
\item \textbf{Supplementary Figure S2}: The sharding workflow begins by downloading authentication tokens, followed by retrieving the package index. After downloading the required package shards, the system determines dependencies from repository metadata, which may trigger additional shard downloads as needed.
\end{itemize}

\subsubsection*{Other Supplementary Materials for this manuscript:}
\textbf{Text Sections:}
\begin{itemize}
\item \textbf{Technical Limitations of Package Management Systems}: Detailed analysis of fundamental limitations in current dependency resolution approaches including SAT solver-based systems (conda/libsolv), backtracking resolvers (pip/PyPI), and heuristic-based systems, with empirical evidence from large-scale studies of package ecosystem failures.
\item \textbf{Pixi's Advanced Dependency Resolution}: Technical implementation details of Pixi's Rust-based SAT solver, including conflict-driven clause learning, and 2-literal watching techniques, constraint handling for mutual exclusion and dependency requirements, and performance benchmarking methodology.
\end{itemize}

\subsection*{Technical Limitations of Package Management Systems}
Technical analysis of existing package management systems reveals fundamental limitations that impact reproducibility. Current dependency resolution approaches fall into three categories, each with distinct limitations: SAT solver-based systems (conda/libsolv), backtracking resolvers (pip/PyPI), and heuristic-based systems (APT).

SAT solver-based systems like conda translate dependency constraints into Boolean satisfiability problems, achieving sophisticated handling of build variants and channel priorities. However, the size of the conda-forge repository makes this a challenging task for regular SAT solvers. Mamba and micromamba, predecessors to pixi, replaced conda's original solver with libsolv, a specialized solver for package management. Yet libsolv itself presents limitations: the code is expert-level C that implements many low-level performance tricks, making it difficult to read and modify. Additionally, libsolv's design is not thread-safe, preventing multi-threaded solving use cases.

Backtracking resolvers suffer from non-deterministic resolution order and excessive backtracking on complex dependency graphs. The Python ecosystem exemplifies these challenges. Peng et al.'s empirical study of 183,864 PyPI library releases found that 35\% suffer from configuration issues, with 68\% requiring source-level analysis rather than version-level checks~\cite{peng2023less}. Their analysis revealed that traditional version-based dependency resolution proves insufficient for real-world reproducibility requirements.

\subsection*{Pixi's Advanced Dependency Resolution}
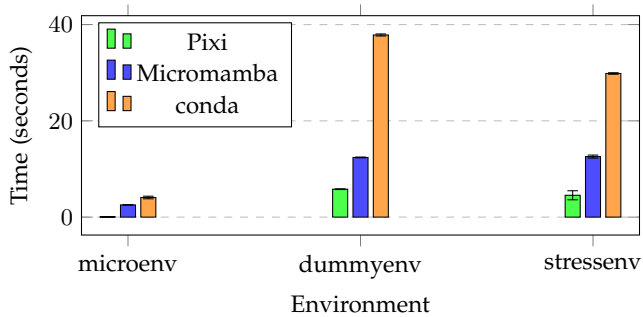
\begin{figure}[th]
\centering
\begin{tikzpicture}
\small
\begin{axis}[
    ybar,
    bar width=0.2cm,
    width=9cm,
    height=4.5cm,
    xlabel={Environment},
    ylabel={Time (seconds)},
    symbolic x coords={microenv, dummyenv, stressenv},
    xtick=data,
    legend pos=north west,
    ymajorgrids=true,
    grid style=dashed,
]

\addplot[fill=green!70, error bars/.cd, y dir=both, y explicit]
coordinates {
    (microenv, 0.0749) +- (0, 0.001)
    (dummyenv, 5.803) +- (0, 0.075)
    (stressenv, 4.539) +- (0, 0.938)
};
\addlegendentry{Pixi}

\addplot[fill=blue!70, error bars/.cd, y dir=both, y explicit]
coordinates {
    (microenv, 2.5384) +- (0, 0.009)
    (dummyenv, 12.396) +- (0, 0.076)
    (stressenv, 12.577) +- (0, 0.310)
};
\addlegendentry{Micromamba}

\addplot[fill=orange!70, error bars/.cd, y dir=both, y explicit]
coordinates {
    (microenv, 4.0756) +- (0, 0.2711)
    (dummyenv, 37.838) +- (0, 0.211)
    (stressenv, 29.840) +- (0, 0.153)
};
\addlegendentry{conda}

\end{axis}
\end{tikzpicture}
\caption{Package manager performance comparison across different environments. Error bars show standard deviation. The microenv installs \texttt{bat}, a \texttt{cat} clone with syntax highlighting and Git integration. The dummyenv installs various visualization libraries, i.e.~\texttt{scipy}, \texttt{polars}, \texttt{ipywidgets}, \texttt{seaborn}, \texttt{matplotlib}, \texttt{plotly}, and \texttt{pandoc} (via \texttt{pip}). The stressenv installs \texttt{rust}, \texttt{jupyterlab} and \texttt{pytorch}.}
\label{fig:package_manager_benchmark}
\end{figure}

Package dependency resolution in scientific computing environments represents a complex constraint satisfaction problem that traditional solvers struggle to handle efficiently. As discussed in the Related Works section, conda's original solver became a severe performance bottleneck as the conda-forge ecosystem grew, with resolution times commonly reaching minutes and occasionally hours for complex scientific software stacks.

Pixi implements a new SAT solver specifically designed for package management, developed in Rust to address fundamental limitations of existing solutions. The libsolv library, while providing significant performance improvements for mamba, presents several constraints that limit its applicability to modern scientific computing requirements: expert-level C implementation that hinders modification and contribution, lack of thread safety preventing parallel solving operations, and maintenance challenges that slow incorporation of improvements.

Pixi's Rust-based solver implementation incorporates established SAT solving techniques including conflict-driven clause learning~\cite{marques2021conflict} and 2-literal watching~\cite{zhang2003cache} for rapid decision making. The solver addresses two primary constraint classes: mutual exclusion rules ensuring single package versions and dependency requirements linking package installations. Unlike general SAT solvers that seek any valid solution, the package management domain requires optimization for maximum version selection, achieved through version-sorted priority testing.

Performance benchmarks demonstrate the practical impact of these improvements. Environment resolution achieves 3× and 10× speedup compared to micromamba and conda, respectively, for typical scientific workloads (Figure~\ref{fig:package_manager_benchmark}). The Rust implementation's native asynchronous capabilities enable parallel downloading, resolution, and installation operations that further accelerate workflow completion.

\subsection*{Sharded Repository Infrastructure}
\begin{figure}
    \centering

\tikzstyle{startstop} = [circle, minimum width=0.4cm, minimum height=0.4cm, text centered, draw=black, fill=black]
\tikzstyle{process} = [rectangle, minimum width=2cm, minimum height=0.8cm, text centered, text width=1.8cm, draw=black, rounded corners, font=\footnotesize]
\tikzstyle{arrow} = [->,>=stealth]

\begin{tikzpicture}[node distance=2.5cm]

\node (start) [startstop] {};
\node (token) [process, right of=start] {download token file};
\node (index) [process, right of=token] {download index file};
\node (shards) [process, right of=index] {download shards for needed packages};
\node (deps) [process, below of=shards, yshift=0.5cm] {determine dependencies from repodata};

\draw [arrow] (start) -- (token);
\draw [arrow] (token) -- (index);
\draw [arrow] (index) -- (shards);
\draw [arrow] ([xshift=2mm]shards.south) -- ([xshift=2mm]deps.north);
\draw [arrow] ([xshift=-2mm]deps.north) -- ([xshift=-2mm]shards.south);

\end{tikzpicture}

    \caption{The sharding workflow begins by downloading authentication tokens, followed by retrieving the package index. After downloading the required package shards, the system determines dependencies from repository metadata, which may trigger additional shard downloads as needed.}
    \label{fig:sharding}
\end{figure}
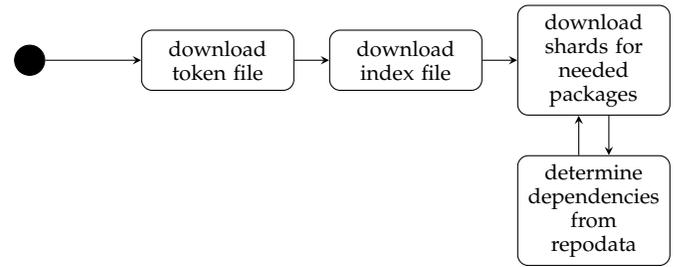

Traditional conda repository management relies on monolithic \texttt{repodata.json} files that encode complete package information for each supported platform. This approach creates scalability bottlenecks as repositories grow: conda-forge now contains approximately 1.5 million packages across 25,000 unique package names, resulting in repodata files reaching hundreds of megabytes.

The monolithic structure forces complete file redownloads whenever packages are added, leading repositories to batch updates hourly for infrastructure efficiency. This batching introduces delays between package publication and availability that impede rapid development cycles common in scientific research.

One way to reduce the time it takes to acquire repodata is to compress the repodata.json files during HTTP transport. When the HTTP server would serve a repodata.json file, it would encode the file on the fly using gzip. This reduces the overall download to around 15\%-30\% of the original size. However, HTTP servers would be limited by the compression speed which reduces the overall transfer speeds to 5MB/s. On high bandwidth connections, this can be a substantial bottleneck.

Pixi introduces a sharded repository architecture that addresses these scalability challenges through content-addressable storage principles. The system decomposes monolithic repodata into an index file containing package name mappings to SHA256-identified shards, with individual shards containing version-specific package information.

This architecture provides several advantages for scientific computing workflows. Individual shards become immutable and cacheable indefinitely due to content addressing, eliminating unnecessary server communication for unchanged packages. The index file scales with unique package counts rather than total package versions, dramatically reducing update overhead. Servers can update immediately upon package publication rather than requiring batched updates, enabling rapid iteration cycles essential for active research projects.

The sharded repodata implementation in Pixi leverages msgpack encoding and zstandard compression for the index file, while individual shards benefit from aggressive CDN caching. The resulting system achieves up to 50× performance improvements for complex environment resolution while reducing infrastructure costs through improved cache utilization.

\end{document}

%% file: abstract.tex
The reproducibility crisis in scientific computing limits the pace of robotics research. Up to 70\% of robotics algorithms cannot be reproduced by independent teams, and many others never release their accompanying code or fail to reach deployment because creating shareable software environments remains prohibitively complex. Although these are distinct challenges, reproducing others’ work and deploying one’s own, they share the same root cause: fragmented, multi-language, and hardware–software toolchains that lead to \emph{dependency hell} and make it difficult to recreate exact computational environments. 
We introduce Pixi, a unified package-management framework that transforms how robotics researchers develop, share, and build upon each other’s work. By capturing exact dependency states in comprehensive project-level lockfiles, Pixi ensures that every installation reproduces an identical environment across platforms. Its Boolean satisfiability (SAT) solver achieves up to 10× faster dependency resolution than comparable package managers, while seamless integration of the conda-forge and PyPI ecosystems eliminates the need to navigate multiple toolchains. 
Since its public release in 2023, Pixi has been adopted in over 5,300 projects, reflecting the growing demand for unified software infrastructure. Our case studies spanning academic collaboration, cross-platform research workflows, field-robotics deployment, and industrial production systems show that Pixi reduces environment setup from hours to minutes while ensuring bit-for-bit reproducibility. Pixi broadens participation in robotics and AI by lowering technical barriers for researchers worldwide, including those in emerging labs and regions. By enabling new forms of collaborative research, from hardware-software integration on desktop machines to large-scale algorithmic benchmarking on high-performance computers, Pixi lays the foundation for the scalable, reproducible infrastructure needed to accelerate future robotics research.